\documentclass[10pt,journal,compsoc]{IEEEtran}

\ifCLASSOPTIONcompsoc
  \usepackage[nocompress]{cite}
\else
  \usepackage{cite}
\fi

\ifCLASSINFOpdf
\else
\fi

\usepackage{subfigure}

\usepackage{times}
\usepackage{epsfig}
\usepackage{hyperref}
\usepackage{graphicx}
\usepackage{amsmath}
\usepackage{amssymb}
\usepackage{boldline, multirow}
\usepackage{xfrac}
\usepackage{tabularx}
\usepackage{booktabs}
\usepackage{pifont}
\usepackage{amsmath, bm}
\usepackage{algorithm,algpseudocode}
\usepackage{xcolor,colortbl}
\usepackage{color, colortbl}

\hypersetup{hidelinks}

\newcommand{\cmark}{$\surd$}
\newcommand{\xmark}{$\times$}
\newcommand{\ie}{\textit{i.e., }}
\newcommand{\eg}{\textit{e.g., }}

\definecolor{LightCyan}{rgb}{0.88,1,1}  
\definecolor{LightRed}{rgb}{1,0.88,0.95}

\begin{document}

\title{Vertical Layering of Quantized Neural Networks for Heterogeneous Inference}

\author{Hai Wu,~ Ruifei He,~ Haoru Tan,~ Xiaojuan Qi,~ Kaibin Huang,~\IEEEmembership{Fellow,~IEEE} 
\IEEEcompsocitemizethanks{
\IEEEcompsocthanksitem Hai Wu, Ruifei He, Xiaojuan Qi and Kaibin Huang are with the Department of Electrical and Electronic Engineering at The University of Hong Kong, Hong Kong.
}}

\IEEEtitleabstractindextext{
\begin{abstract}
Although considerable progress has been obtained in neural network quantization for efficient inference, existing methods are not scalable to heterogeneous devices as one dedicated model needs to be trained, transmitted, and stored for one specific hardware setting, incurring considerable costs in model training and maintenance. 
In this paper, we study a new vertical-layered representation of neural network weights for encapsulating all quantized models into a single one. 
It represents weights as a group of bits (\ie vertical layers) organized from the most significant bit (also called the basic layer) to less significant bits (\ie enhance layers).
Hence, a neural network with an arbitrary quantization precision can be obtained by adding corresponding enhance layers to the basic layer. 
With this representation, we can theoretically achieve any precision network for on-demand service while only needing to train and maintain one model.
However, we empirically find that models obtained with existing quantization methods suffer severe performance degradation if adapted to vertical-layered weight representation. 
A significant issue is that once a quantization model is well-trained for a certain precision, it is difficult to reuse the weights and transfer them to another precision since essential information lying in the full precision source model is discarded after dedicated training.
To this end, we propose a simple once quantization-aware training (QAT) scheme for obtaining high-performance vertical-layered models. 
Our design incorporates a cascade downsampling mechanism which allows us to obtain multiple quantized networks from one full precision source model by progressively mapping the higher precision weights to their adjacent lower precision counterparts.
The mapping is achieved by discarding the least significant bits of the higher precision model. 
Then, with networks of different bit-widths from one source model, multi-objective optimization is employed to train the shared source model weights such that they can be updated simultaneously, considering the performance of all networks.
By doing this, the shared weights will be optimized to balance the performance of different quantized models, thus making the weights transferable among different bit widths. 
After the model is trained, to construct a vertical-layered network, the lowest bit-width quantized weights become the basic layer, and every bit dropped along the downsampling process act as an enhance layer. 
Our design is extensively evaluated on CIFAR-100 and ImageNet datasets. Experiments show that the proposed vertical-layered representation and developed once QAT scheme are effective in embodying multiple quantized networks into a single one and allow one-time training, and it delivers comparable performance as that of quantized models tailored to any specific bit-width. Code will be available. 
\end{abstract}

}

\maketitle
\IEEEdisplaynontitleabstractindextext
\IEEEpeerreviewmaketitle

\IEEEraisesectionheading{
\section{Introduction}
\label{sec: introduction}
}

\IEEEPARstart{D}{eep} neural networks (DNNs) have achieved revolutionary progress in many intelligent services, including image recognition, video segmentation, machine translation, and autonomous driving. However, the large storage footprint and heavy computation cost hinder the deployment of DNNs on edge devices with computation constraints, and accelerating DNNs has therefore become a significant research problem.

\begin{figure}[t]
    \begin{center}
    \includegraphics[width=1\linewidth]{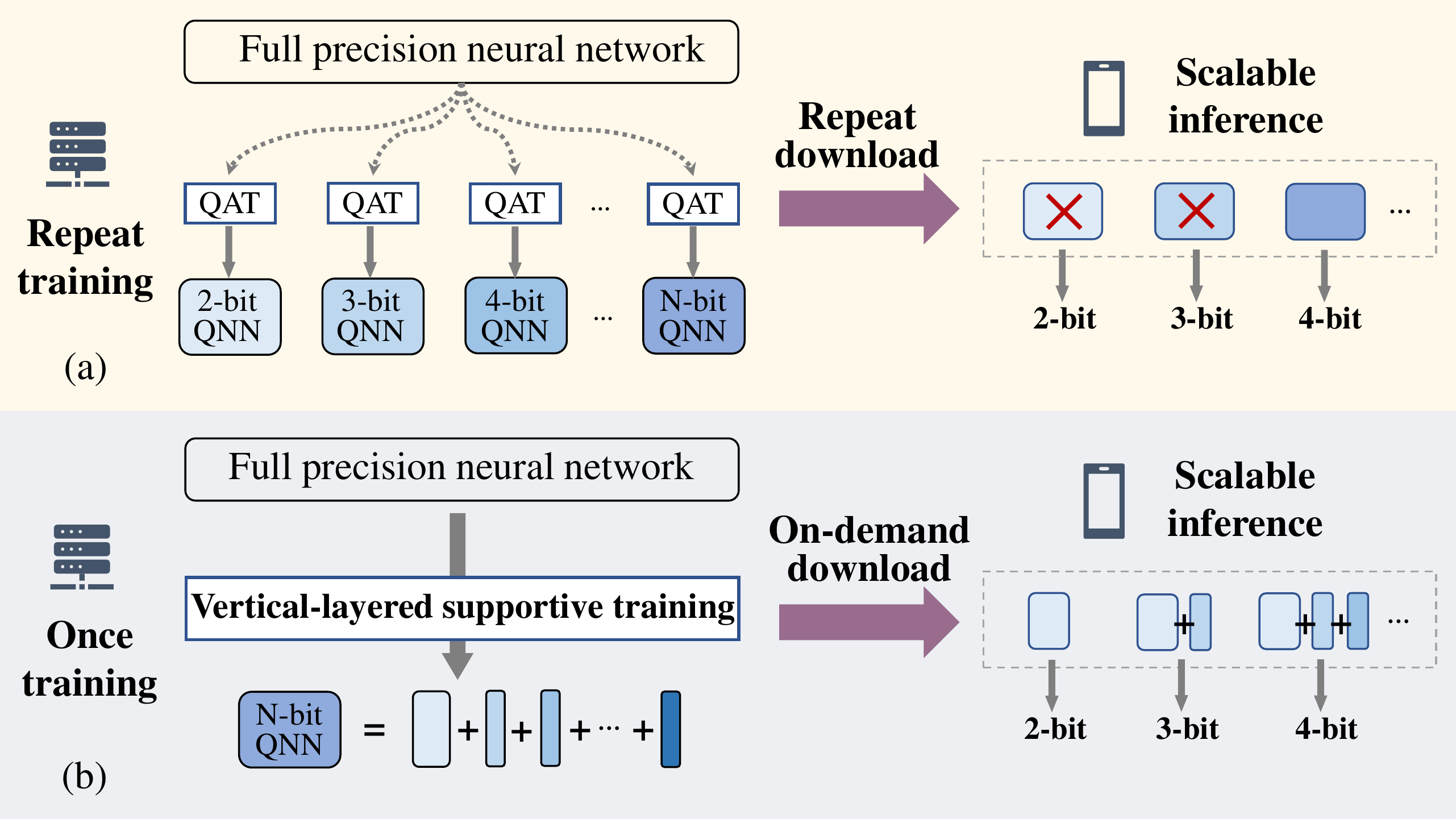}
    \end{center}
    \caption{For heterogeneous deployments: (a) existing quantization frameworks have to conduct quantization-aware training and store one separate model for each specific setting;
    and (b) our model with vertical-layered weight representation allows one-time training and encapsulates all quantized models into a single one for on-demand deployment.}
    \label{fig: scalable inference comparison}
\end{figure}

Among existing model acceleration techniques, neural network quantization is considered a promising direction to shrinking model size and making inference efficient by reducing the number of bits involved in the calculation, transmission, and storage \cite{DC_Han, CDNN_Gong, QNN_Itay}. 
The quantized networks are often obtained via a dedicated quantization-aware training (QAT) process \cite{DoReFa_Zhou, QAT_Park, LQNet_Zhang, PACT_Choi, DSQ_Gong, LSQ_Steven}, where the quantized parameters are used in the forward pass to mimic the inference process, and the gradient approximated by straight-through-estimator \cite{STE_Krizhevsky} is obtained in the backward pass to update the full precision source model parameters.
Although promising results have been attained, existing schemes only allow obtaining one network of a specific precision at a time. 
To serve heterogeneous devices, many dedicated models have to be designed, trained, and stored for diverse hardware settings. This incurs huge computational costs in model training and requires ample storage space to save all quantized models, making existing schemes not scalable to heterogeneous settings; see Fig.~\ref{fig: scalable inference comparison}(a).

To this end, we propose a new \textit{vertical-layered representation} of neural network weights for encapsulating all quantization models of different precisions into a single one. 
It represents weights as a group of bits, which are coined as \textbf{vertical layers}, and organizes them from the most significant bit (basic layer) to a series of less significant ones (enhance layers). 
Each vertical layer represents one quantization resolution and is expected to be compatible with the others for assembling different bit-width quantization networks. 
Hence a neural network with the desired quantization setting can be obtained by adding a number of enhance layers to the basic layer model.
With this representation, we only need to train one vertical-layered model and store one copy of the parameters, which can be used to assemble any precision quantized network to support heterogeneous inference; see Fig.~\ref{fig: scalable inference comparison}(b).
Albeit conceptually appealing, obtaining a vertical-layered neural network remains a challenging problem. Our initial investigation shows that the performance of existing methods degrades dramatically if the models are treated as a vertical-layered network directly; see Sec.~\ref{subsec: naive approaches}.

The major obstacle to developing such vertical-layered networks lies in the diverse weight distribution tailored to different bit-width settings and difficulties in optimizing the model involving multiple groups of discrete parameters simultaneously. 
As illustrated in Fig.~\ref{fig: weight histogram comparison}(a), to achieve the best performance for all precision levels, the distribution and range of weights vary in different bit-width settings: they concentrate around zero-point and have a narrow range for lower precision networks (\eg 2-bit) but tend to show a long-tail distribution in high bit-width settings (\eg 4-bit). 
This makes network weights optimized to a particular precision level challenging to be reused by another precision level, which causes performance degradation; see Fig.~\ref{fig: weight histogram comparison}(b).
Besides, joint training of discrete model parameters using the existing method to make them fit all precision levels suffers from the conflict requirements of networks with different bit widths, which is unfolded in Sec.~\ref{subsec: naive approaches}.
Therefore, a new QAT scheme needs to be carefully re-designed to alleviate interference among vertical-layered representations and tackle the challenges in optimizing multiple quantization networks, especially at extremely low bit-width settings.

To overcome the challenges above, we propose a once QAT scheme for obtaining a desired vertical-layered network without sacrificing model performance too much. 
First, to support vertical-layered representation and alleviate conflicts among networks of different bit widths, we introduce a cascade downsampling mechanism that directly maps weights of a relatively higher precision level to their adjacent lower precision counterparts by dropping the least significant bits residing in the former ones. 
Hence, based on one full precision source model, we can have multiple quantization networks of different precisions by progressively reducing its bit widths. 
Second, given multiple quantized networks from one full precision source model, we optimize the underlying source parameters by minimizing an overall loss of all quantized networks and employ multi-objective optimization (MOO) to balance the contribution of each quantized model to the overall loss. 
Through such training, the shared source parameters will be updated to balance the performance of all quantized networks and make the weights transferable among networks of different precisions; see Fig.~\ref{fig: weight histogram comparison}(c). 
Third, as our design naturally encloses multiple networks with diverse capabilities, we study a self-knowledge distillation paradigm where the lower precision network can get supervised by the higher one to strengthen its performance instead of resorting to extra teacher models.
Finally, after our source model is trained, to construct a vertical-layered network, the lowest bit-width quantized weights become the basic layer, and every bit dropped in the downsampling process acts as an enhance layer.

We conduct extensive experiments on CIFAR-100 \cite{CIFAR_Krizhevsky} and ImageNet datasets \cite{ImageNet_Russakovsky} to verify the vertical-layered weight representation and assess the effectiveness of our proposed design. 
To our best knowledge, without resorting to computationally heavy neural architecture search methods, we are the first to demonstrate a high-performance vertical-layered low-bit-width model which embeds multiple quantized networks into a single one and allows one-time training.
At the same time, for all evaluated bit widths, our model can deliver comparable performance as that of models optimized for a specific bit width. Notably, our model upon ResNet \cite{ResNet_He} backbone can even outperform existing methods that need QAT for each bit width in some settings. 
Further, we extend the uniform-precision trained model via MOO to the mixed-precision setting by sampling a series of quantized weights from our model via minimizing the weight quantization error given a total network size budget. 
We obtain performance improvements as the number of bit budgets increases, which manifests the scalability of our method in incorporating many quantized models into a single one.

The remaining of this paper is organized as follows. Related works are summarized in Sec.~\ref{sec: related work}. Then we revisit neural network quantization and introduce the concept of vertical layering in Sec.~\ref{sec: quantization and representation}. We elaborate our design for training vertical-layered representation models in Sec.~\ref{sec: our method} and make a detailed discussion in Sec.~\ref{Sec: discussion}. We conduct experiments and present concise analysis in Sec.~\ref{sec: experiments}. Finally, we draw our conclusion in Sec.~\ref{Sec: conclusion}.

\section{Related Work} 
\label{sec: related work}

\subsection{Neural Network Quantization}
\label{subsec: neural network quantization}

Neural network quantization attempts to approximate the full precision data using a fixed set of code words for saving bits \cite{DC_Han}, thus slimming the model size and converting the general floating-point arithmetic into low-complexity integer operations. 
Based on training schemes, existing methods can be categorized into two major streams: the post-training quantization (PTQ) scheme and the quantization-aware training (QAT) technique. 
Early works adopted PTQ of DNNs \cite{Accurate_PTQ, PTQ_Wang, PTQ_Banner, PTQ_Nahshan, PPLQ_Fang}, where the mapping from full precision weights and activations to fixed integers are optimized after the traditional network training. PTQ requires no (or only a small amount of) training data for optimization and achieves nearly no accuracy drop in \textit{int8} quantization. Yet, it generally suffers significant degradation when shrinking the quantization bit-width further \cite{PTQ_Wang}. 
Another group of works \cite{DoReFa_Zhou, QNN_Itay, PACT_Choi, LQNet_Zhang, LSQ_Steven}, \ie QAT, incorporates the quantization process along with the network training to optimize the network weights for a more compact quantization with better performance. 
QAT inserts the weight and activation discretization process in the network forward pass and uses the straight-through estimator \cite{STE_Krizhevsky} to approximate the gradient for updating original full precision parameters, which generally needs dedicated optimization. Two extreme cases, binary neural networks (BNNs) \cite{BNN_Cour, BiRealNet_Liu, Real2Bin_Brais, HWGQ_Cai} and XOR-Net \cite{XOR_Rast, TQ_Zhu}, use only binary or ternary representations of weights or activations to achieve single bit forward inference, which immensely reduces the arithmetic calculation at the cost of significant accuracy degradation. 

Existing schemes can also be categorized into uniform-precision quantization and mixed-precision quantization based on the uniformity of quantization precision. 
The uniform-precision quantization requires different layers within one network to share the same bit-width settings \cite{Accurate_PTQ, DoReFa_Zhou, QNN_Itay}. 
The mixed-precision quantization scheme allows the usage of different bit widths for different layers within one network \cite{HAQ_Wang}, which aims to find a suitable bit allocation scheme for each layer of the network model to minimize the accuracy degradation within the limited bit operation/storage budget. 
Compared with uniform-precision quantization, mixed-precision quantization can achieve better deployment flexibility and good trade-offs in performance and efficiency. 
Therefore, it has received extensive attention from the academic community \cite{HAQ_Wang, HAWQ_Dong, HAWQ_V3_Yao}. 
Since mixed-precision quantization often requires complex hardware adaptation, practitioners often choose the hardware-friendly uniform-precision quantization technology in practice.

However, the above works focus on training one network with a certain precision at one time, which is not scalable for heterogeneous application scenarios. 
In this paper, we focus on developing a vertical-layered representation of neural network weights to embed multiple neural networks into one model, offering one-time training to serve on-demand deployment.
To that end, we develop new QAT schemes. 
Besides, although our primary analysis is conducted on uniform-precision quantization, we further show the flexibility of our method by extending it to the mixed-precision scenario.

\subsection{Dynamic Neural Networks}
\label{subsec: dynamic neural networks}

Dynamic neural networks adjust architecture or parameters to satisfy different computation requirements while maximizing model performance.
Architecture-oriented dynamic networks often change the computation graph by selectively activating a subset of model components \cite{Dynamic_NN}.
Adaptive neural network \cite{AdaptiveNN_Boluk} introduces early-exit branches at different depth levels such that the model can generate outputs at intermediate layers. 
Further, some other works \cite{SkipNet_Wang, BlockDrop_Wu, CGNet_Hua} add gating modules or skip controllers to drop layers or channels during inference.
Unlike the above approaches realizing dynamic depth inference, the slimmable network \cite{Slimmable_Yu} can execute at different layer-wise widths, \ie the number of active channels. 
Recently, neural architecture search has become another paradigm for achieving dynamic networks, which generally trains a superfluously parameterized network (\ie supernet), and a desirable sub-network can be retrieved through sampling and transformation \cite{NAS_PS_Pham, NAS_SP_Guo, NASS_Li}. 
Based on the once-for-all network \cite{OFA_Han}, which firstly enlarges the architecture search space to obtain promising elastic models to suit varied deployment scenarios, the work in \cite{APQ_Wang} incorporates the mixed precision selection and pruning for parameters in the searching stages to achieve further flexibility. 
These techniques provide a satisfactory network architecture under given resource constraints, which is orthogonal with our proposed vertical-layered network representation.

The work most related to ours is the Adabits \cite{AdaBits_Jin}, which achieves adaptive bit-width inference based on QAT. 
However, it only achieves satisfactory accuracy in high bit-width settings (4-bit and above). It is difficult for the model to obtain a low-bit-width model suited for vertical-layered representation, which requires no access to the full precision source parameter for scalable deployment.
Meanwhile, existing works demonstrate that a 4-bit quantization network is already sufficient to achieve the performance of a full precision network baseline model \cite{PACT_Choi} \cite{LSQ_Steven}, which undermines the applicability of this work.
In contrast, we focus on more challenging low-bit-width scenarios and demonstrate that our developed vertical-layered model achieves favourable performance at an extremely low-bit-width setting (\ie 2-bit) while maintaining dynamic flexibility.

\subsection{Multi-Objective Optimization}
\label{subsec: multi-objective optimization}

Multi-objective optimization (MOO) involves simultaneously optimizing a set of possibly conflicting objectives, which is popularly adopted in multi-task learning \cite{MTL_MOO_Sener, NL_MOO_Miettinen}. 
One widespread practice is performing a weighted sum of all objective functions to transform the original problem into a single-objective problem.
The prior knowledge of the combination weights is critical for achieving the optimum, \ie the Pareto optimal solution \cite{SurveyMOO_Hwang}. 
Then, the gradient-based optimization methods are developed to iteratively update the combination weights by solving a convex combination problem, which provides a descent direction to arrive at the Pareto front by analyzing the multi-objective Karush-Kuhn-Tucker (KKT) conditions \cite{SDM_MOO_Fliege, SQP_MOO_Fliege, MGDA_Desideri}.
Recent work \cite{US_Kurin} demonstrates that the different MOO optimizers act as regularizers in the original MOO problem, and the unitary scalarization is capable of achieving the optimal solutions.
Since our vertical-layered model embodies multiple networks that can yield multiple outputs, the training process must be carefully coordinated to balance different objectives. 
In this work, we formulate the training of a vertical-layered supportive network as a MOO problem and investigate different MOO optimizers to obtain a high-performance vertical-layered quantization model.

\subsection{Knowledge Distillation}
\label{subsec: knowledge distillation}

Knowledge distillation transfers the knowledge learned by large models (\ie teacher) to a compact one (\ie student) via a weighted combination of original loss using ground-truth labels and distillation loss calculated with the soft output of a teacher model \cite{KD_Geoffrey}. 
There is a broad consensus that knowledge distillation can help improve the performance of low-bit-width quantized networks by transferring the knowledge of a full precision network to quantized ones \cite{Apprentice_Mishra, QKD_Kim, SKDQ_Boo}. 
QKD \cite{QKD_Kim} proposes a quantization-aware knowledge distillation scheme by co-training a low-bit-width student and a larger teacher model. 
SPEQ \cite{SKDQ_Boo} introduces a novel self-training paradigm where the high-precision teacher and the low-bit-width student from the same full precision source model are trained collaboratively. 
In our training process, where different bit-width networks are enclosed, we design a self-knowledge transfer that utilizes a relatively higher precision network's output as soft labels for the cooperative training of the lower precision ones.

\section{Quantization and Vertical-Layered Representation}
\label{sec: quantization and representation}

In the following, we first revisit the quantization of neural networks in Sec.~\ref{subsec: quantization}. Then in Sec.~\ref{subsec: vertical layering of network}, we elaborate on the proposed vertical-layered representation and demonstrate how to decompose the weight parameters into multiple vertical layers. Furthermore, in Sec.~\ref{subsec: naive approaches}, we provide some baseline methods and present the challenge of obtaining vertical-layered models.

\subsection{Quantization}
\label{subsec: quantization}

Without loss of generality, we consider the uniform quantization process \footnote{It is necessary to distinguish between \textit{uniform quantization} and \textit{uniform-precision quantization}, where uniform quantization refers to the type of quantization in which the quantized levels are uniformly spaced, while uniform-precision quantization denotes that all layers within a network share the same bit-width settings. The non-uniform quantization generally requires a specific hardware design and is not favored in general heterogeneous applications.}.
Given a real-valued tensor $\bm{v}$, which can be weight parameters $\bm{w}$ or input activation $\bm{a}$, the quantization process $Q(\cdot, \cdot)$ and de-quantization process $\hat{Q}(\cdot, \cdot)$ can be formulated as:
\begin{equation}
\begin{aligned}
    \bar{\bm{v}} & = Q(\bm{v}, {s}) = \mathrm{{clip}} ( \lfloor \frac{\bm{v}}{{s}} \rceil, -Q_N, Q_P), \\
    \hat{\bm{v}} & = \hat{Q}(\bar{\bm{v}}, s) = \bar{\bm{v}} \times s;
\end{aligned}
\label{eq: uniform quantization}
\end{equation}
where $\bar{\bm{v}}$ and $\hat{\bm{v}}$ denote the quantized tensor and de-quantized tensor respectively, $\lfloor \cdot \rceil$ indicates the rounding operation that maps a continuous number to its nearest integer number, and $\mathrm{clip}(\bm{x}, r_1, r_2)$ returns a tensor with entries below $r_1$ set to $r_1$ and entries above $r_2$ set to $r_2$. 
We use $s \in \mathbb{R}_{+}$ to denote the quantization step size, which determines the smallest resolution of a quantization level. 
Another quantization parameter, $[-Q_N, Q_P]$, represents the quantization range that constrains the smallest quantized integer to be $-Q_N$ and the largest integer to be $Q_P$. 
Given the number of bits $K$, for signed data, $Q_N=2^{K-1}$ and $Q_P=2^{K-1}-1$; for the unsigned case, $Q_N=0$ and $Q_P=2^{K}-1$.

\begin{figure}[t]
    \centering
    \includegraphics[width=0.9\linewidth]{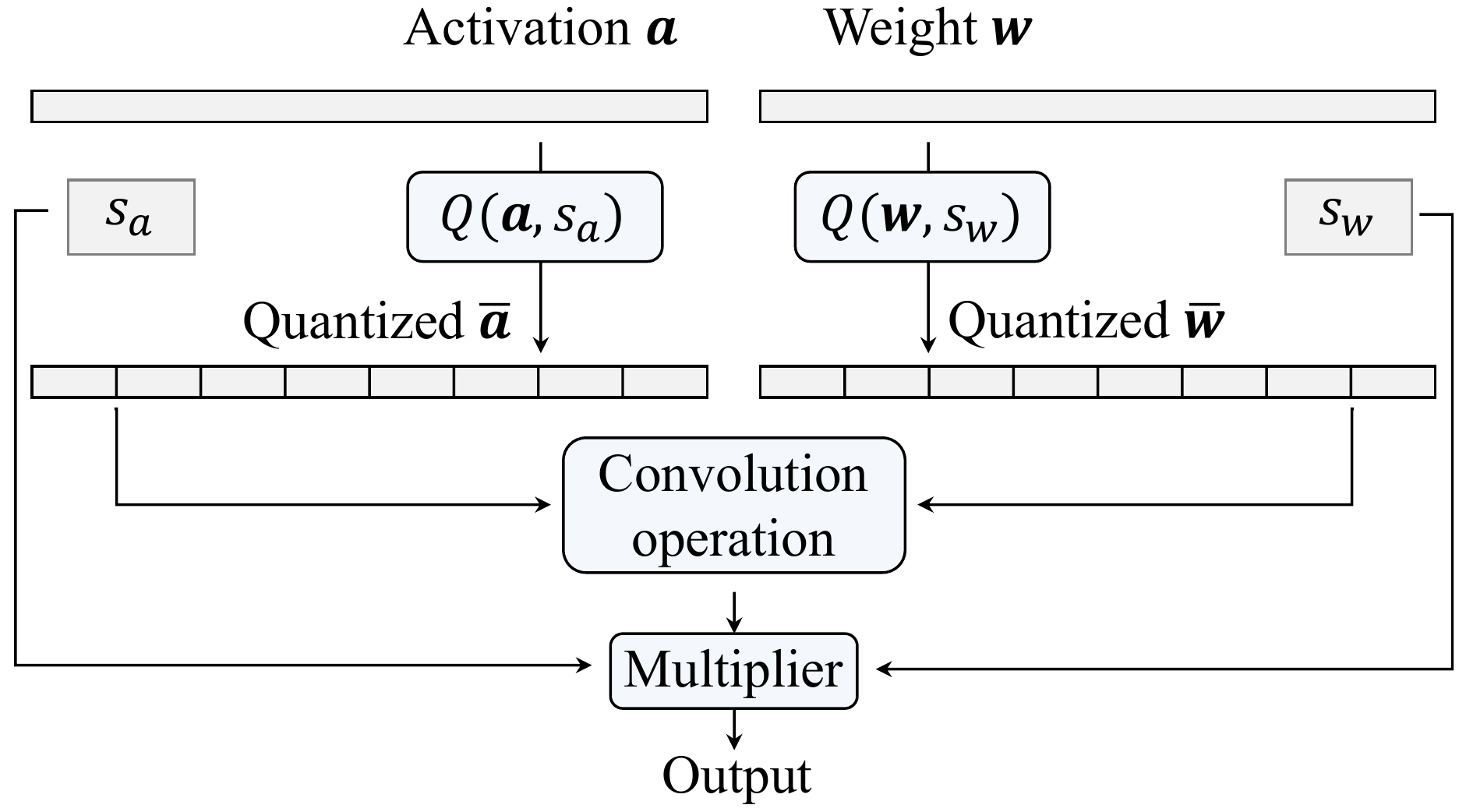}
    \vspace{-0.1cm}
    \caption{The data flow of a quantized convolution operator. The quantization of weight can be done offline, and activation quantization is performed online. The quantized values are represented in low bit-width to simplify the convolution multiplication.}
    \label{fig: LSQ}
\end{figure}

The data flow involved in a quantized convolution operator during the inference stage is illustrated in Fig.~\ref{fig: LSQ}. 
The real-valued weight $\bm{w}$ and activation $\bm{a}$ are quantized into low bit-width representations $\bar{\bm{w}}$ and $\bar{\bm{a}}$ respectively according to Eq.~\eqref{eq: uniform quantization}. Separate step sizes $s_a$ and $s_w$ are used for weight and activation quantization respectively. 
Then, quantized weights $\bar{\bm{w}}$ and activations $\bar{\bm{a}}$ are used to perform the low-precision convolution operation, \ie
\begin{equation}
\begin{aligned}
    \bm{y} &= (\bar{\bm{w}} \otimes \bar{\bm{a}}) \times s_w \times s_a;
\label{eq:conv quant 0}
\end{aligned}
\end{equation}
where $\otimes$ denotes the simplified convolution calculation. 
Finally, the final output $\bm{y}$ is obtained by multiplying these two step sizes on the computed results.

In QAT, the forward phase is calculated using the quantized/de-quantized tensor, while the backward phase aims to update the full precision source parameters. 
Since the discrete rounding and clip operation between the full precision $\bm{v}$ and the quantized $\bar{\bm{v}}$ in Eq.~\eqref{eq: uniform quantization} truncate the gradient propagation path, the gradient estimator (\eg straight-through-estimator \cite{STE_Krizhevsky}) must be adopted for enabling gradient back-propagation from $\bar{\bm{v}}$ to $\bm{v}$. 
The step size $s$ could be set as fixed values as in many existing works \cite{DoReFa_Zhou, PACT_Choi, LQNet_Zhang, Effective_Zhuang}. 
The work in \cite{LSQ_Steven} demonstrates that treating step sizes as learnable parameters and jointly optimizing them with the model parameters will improve model performance. 
In this work, we adopt such a learnable quantization step size for good flexibility and high efficiency.

\begin{figure}[t]
    \centering
    \includegraphics[width=0.95\linewidth]{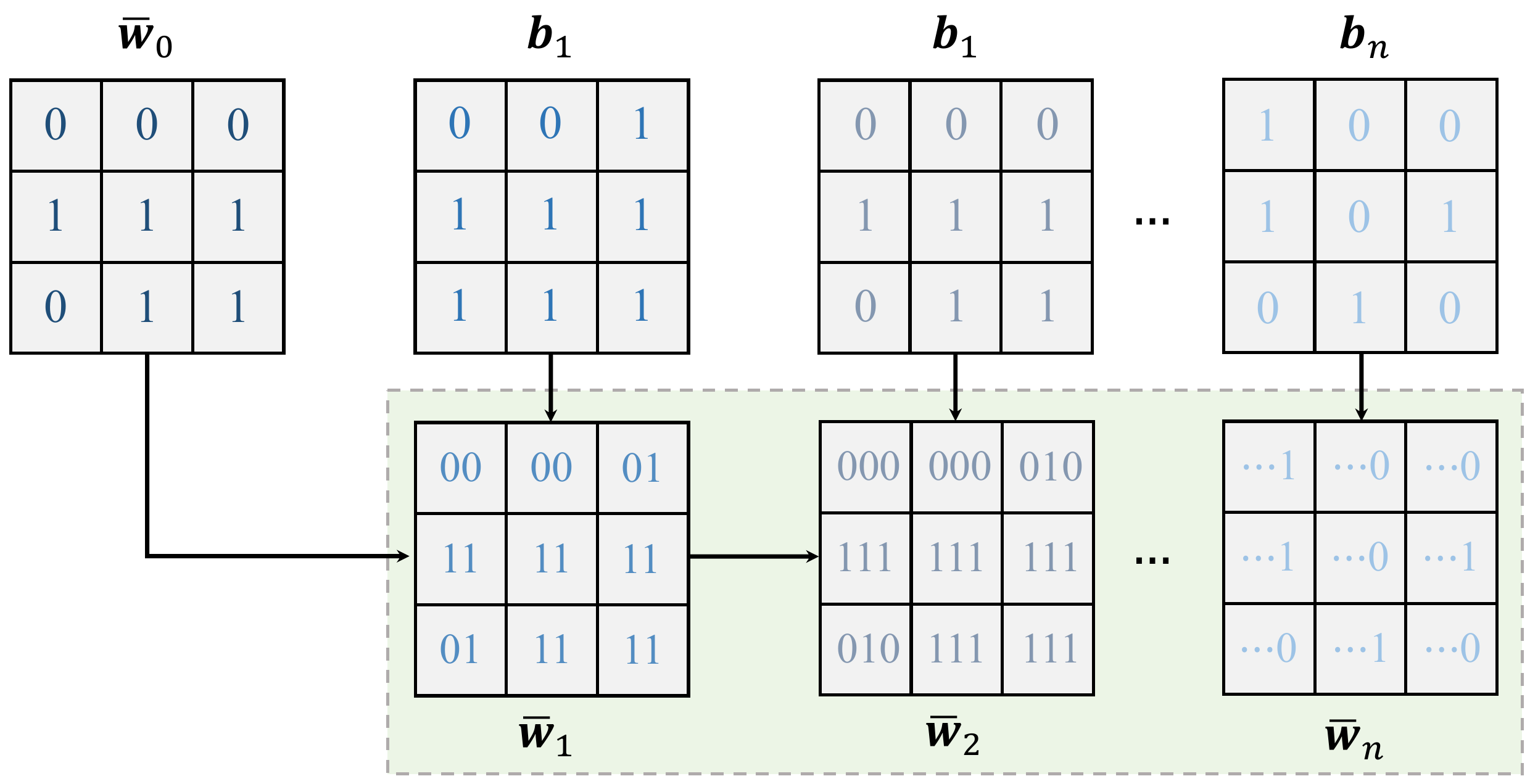}
    \caption{Vertical-layered representation of a 2D $3\times3$ convolution kernel.}
    \label{fig: vertical-layered representation}
\end{figure}

\subsection{Vertical Layering of Network Weights}
\label{subsec: vertical layering of network}

Existing quantization methods consider designing one dedicated model for a specific precision, which is not scalable to practical heterogeneous settings. In the following, we will introduce our proposed simple vertical-layered representation of neural network weights, which allows encompassing multiple quantized networks into a single one.

As shown in Fig.~\ref{fig: vertical-layered representation}, the vertical-layered representation divides the weight parameters into different groups of bits.
The quantized weight parameters, $\bar{\bm{w}}$, are partitioned into a low bit-width tensor $\bar{\bm{w}}_0$ and a series of independent and hierarchical binary tensors $\{\bm{b}_i\}_{i=1}^n$, each one is of the same shape as original $\bm{w}$.
The parameter set $\bar{\bm{w}}_0$ consisting of the most significant bit of weight is termed as the \textit{basic layer}.
And the sequential groups of less significant bits $\{\bm{b}_i\}_{i=1}^n$ are called \textit{enhance layers}. 
One can directly use the basic layer $\bar{\bm{w}}_0$ to construct a quantized neural network running at the lowest precision, and also can build up a higher bit-width quantized network by adding additional enhance layers to $\bar{\bm{w}}_0$ sequentially.
With vertical-layered representation, heterogeneous edge devices can access any desired bit-width quantization network in an on-demand manner (see Fig.~\ref{fig: scalable inference comparison}(b)).

To support the vertical-layered representation of quantized neural networks, the following rule is constructed to build the relationship between quantized weights of different bit widths, 
\begin{equation}
\begin{aligned}
    \bar{\bm{w}}_i &= 2\times \bar{\bm{w}}_{i-1} + \bm{b}_i, \\
    s_i &= \sfrac{1}{2} \times s_{i-1}; \\
    \label{eq: bit-inheritance}
\end{aligned}
\end{equation}
where $\bar{\bm{w}}_{i-1}$ indicates the adjacent lower bit-width quantized weights of the quantized weights $\bar{\bm{w}}_i$. 
The high precision $\bar{\bm{w}}_i$ is constructed by integrating the $i$-th enhance layer $ \bm{b}_i$ with the lower bit-width $\bar{\bm{w}}_{i-1}$. Thus, the subscript $i \in \{1, ..., n\}$ indicates that the quantized weight is assembled by $i$ enhance layers.
This rule ensures that the high-precision quantized weights shall comprise the adjacent lower-precision quantized ones, which guarantees they are compatible with assembling any bit-width quantized networks. 
The recursive rule in Eq.~\eqref{eq: bit-inheritance} could be rewritten as the following form, 
\begin{equation}
    \bar{\bm{w}}_i = 2 ^ i \times (\bar{\bm{w}}_0 + \sum_{j=1}^i 2^{-j} \cdot \bm{b}_j ). 
    \label{eq: linear interpolation}
\end{equation}

\begin{figure}[t!]
    \centering
    \includegraphics[width=0.9\linewidth]{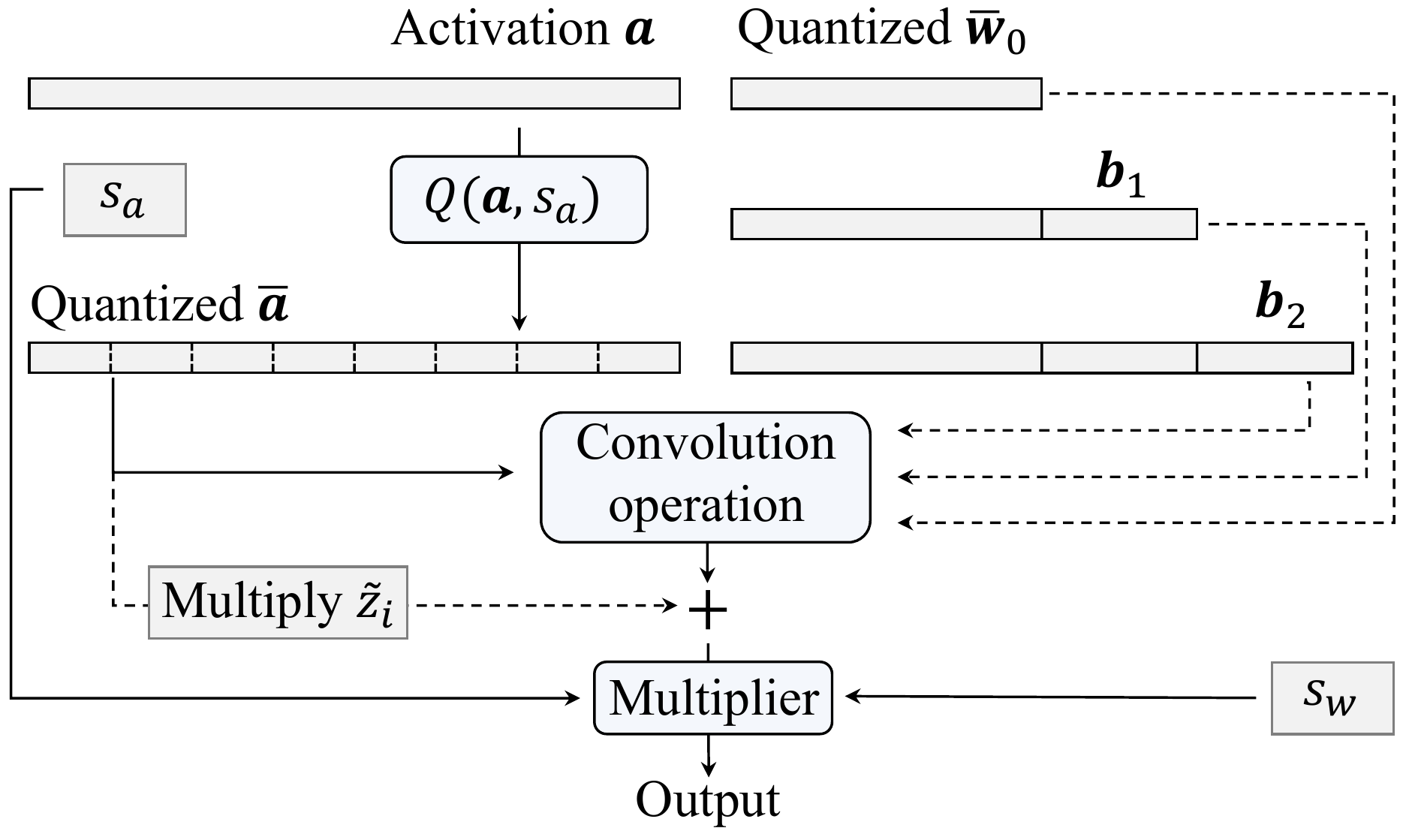}
    \vspace{-0.1cm}
    \caption{Data flow for scalable inference. The quantized weights with any precision can be used forwarding convolution operation. And the input activation is quantized into corresponding bit width to match the computation precision.}
    \label{fig: scalable inference}
\end{figure}

For a better illustration, we show the data flow in the convolution block under the vertical-layered representation setting in Fig.~\ref{fig: scalable inference}. The desired bit-width $\bar{\bm{w}}_i$ is obtained by incrementally adding a number of enhance layers to the basic layer. And the activation tensor $\bm{a}$ is quantized by $Q(\bm{a}, s_a)$ on the fly to achieve the corresponding precision to match the computation requirements. 
Finally, the corresponding step size $s_i$ and $s_a$ are used to scale the convolution results as the final output.

\subsection{Baseline Method and Research Challenges}
\label{subsec: naive approaches}

This subsection discusses the unresolved technical challenges faced by the vertical-layered representation. We introduce two intuitive baseline approaches for obtaining a vertical-layered neural network: the \textit{ascending layer training} and the \textit{multi-layer co-training}. 

\vspace{0.2cm}
\noindent
\textbf{Ascending layer training.} In this approach, the basic layer $\bar{\bm{w}}_0$ and each enhance layer in $\{\bm{b}_i\}_{i=1}^n$ are learned sequentially. 
Firstly, the basic layer $\bar{\bm{w}}_0$ is generated through dedicated QAT as elaborated in Sec.~\ref{subsec: quantization}. 
Then, for training the binary enhance layer $\bm{b}_i$, we introduce another full precision source parameter $\bm{b}_i^r$ and set $\bm{b}_i = \text{sign}(\bm{b}_i^r)$, where the $\text{sign}$ function outputs entry-wise $1/0$ according to the sign of each entry in $\bm{b}_i^r$.
By substituting $\bm{b}_{i}$ into Eq.~\eqref{eq: bit-inheritance}, we have 
$$
\bar{\bm{w}}_{i} = 2 \times \bar{\bm{w}}_{i-1} + \text{sign}(\bm{b}^r_i). 
$$ 
During training, the straight-through-estimator \cite{STE_Krizhevsky} is used to approximate the backward gradient from the binary $\bm{b}_i$ to source parameter $\bm{b}^r_{i}$. 
The training of all enhance layers contains $n$ stages, and the $i$-th stage will learn the $i$-th enhance layer $\bm{b}_i$.

\vspace{0.2cm}
\noindent
\textbf{Multi-layer co-training.} The second baseline is to optimize all vertical layers simultaneously. 
Following the settings in the first baseline, we also introduce the full precision tensor $\bm{b}_i^r$ to learn the binary enhance layer $\bm{b}_i$, where $\bm{b}_i = \text{sign}(\bm{b}_i^r)$ and the gradient is also estimated via the straight-through-estimator \cite{STE_Krizhevsky}. 
With Eq.~\eqref{eq: linear interpolation}, we can decompose any bit-width tensor $\bar{\bm{w}}_i$ into the combination of the basic layer and corresponding vertical layers, 
$$
\bar{\bm{w}}_i = 2 ^ i \times (\bar{\bm{w}}_0 + \sum_{j=1}^i 2^{-j} \cdot \text{sign}(\bm{b}_j^r)). 
$$
During training, networks with different bit-width parameters perform forward and backward passes concurrently; hence all the source parameters of the vertical layers are optimized simultaneously.

\begin{figure}[t]
	\centering	
	\includegraphics[width=1.0\linewidth]{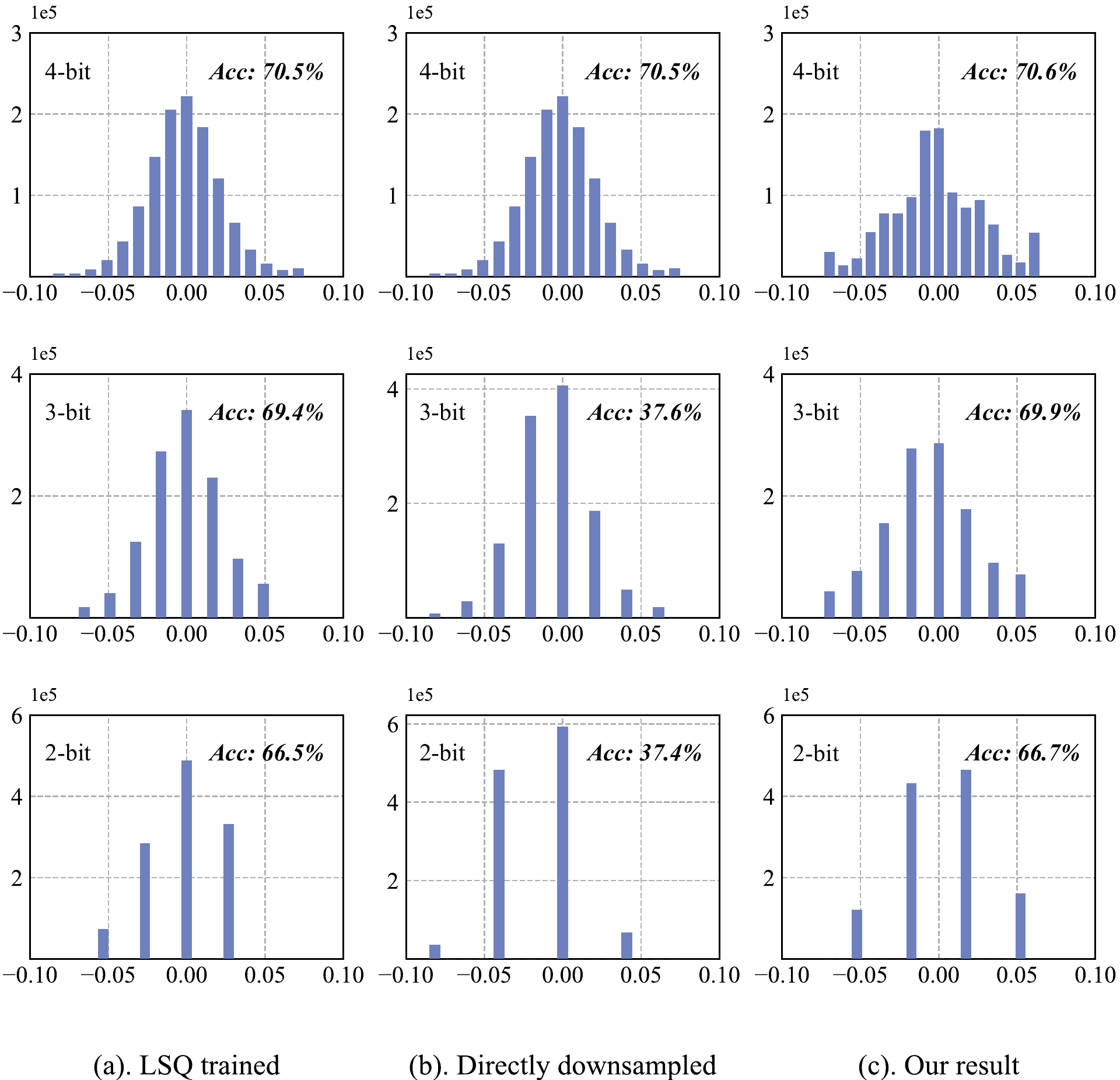}
	\vspace{-0.4cm}
	\caption{Weight histograms of different bit-width quantized weights in ResNet-18 trained on ImageNet dataset based on different approaches, Top-1 validation accuracy is noted. Column (a). Different bit-width quantized networks are trained individually using LSQ \cite{LSQ_Steven} to achieve the best representation capability under specific precision criteria. Column (b). The 4-bit network is obtained through QAT, and the lower bit-width weights are directly downsampled from the highest precision network with fine-tuning. Column (c). Different precision networks are jointly trained once based on our proposed training framework.}
    \label{fig: weight histogram comparison}
\end{figure}

\vspace{0.2cm}
\noindent
\textbf{Challenge analysis.} We empirically evaluate the ascending layer training and multi-layer co-training methods and compare them with the oracle results (obtained by bit-width tailored QAT in \cite{LSQ_Steven}). 
The results are shown in Table~\ref{tab: baseline comp}. 
For the ascending layer training scheme, there is a nearly $2\%$ top-1 accuracy degradation in the 3-bit and 4-bit networks in comparison with the oracle results. 
To investigate the reason, we visualize the weight histogram of the oracle network with different bit widths in Fig.~\ref{fig: weight histogram comparison}(a). 
Obviously, the weight distributions of networks of different bit widths vary a lot, \ie the weights of a 4-bit quantized network show a more long-tailed distribution while the weights of the 2-bit one are more concentrated around zero. 
However, in the ascending layer scheme, the weight distribution of a higher bit-width network highly depends on the weight of the lower bit-width networks as the more significant bits are frozen when the model progressively grows by adding more ascending layers to support the vertical-layered representation. 
As for the second baseline, we find that the training process easily diverges because it involves concurrently optimizing multiple binary parameter sets. 
Therefore, we propose a new joint QAT scheme to achieve high-performance vertical layering supportive models.

\begin{table}[h]
\renewcommand\arraystretch{1.4}
    \centering
    \caption{Baseline approaches investigation. Experiments are conducted using ResNet-18 \cite{ResNet_He} architecture on ImageNet dataset \cite{ImageNet_Russakovsky}. The basic layer is composed of a 2-bit quantized network. W/A indicates the weight and activation are quantized into W-bit and A-bit respectively. Top-1 accuracy is reported.} 
    \begin{small}
    \setlength{\tabcolsep}{4.5mm}{
        \begin{tabular}{l|ccc}
        \bottomrule[1pt]
        \multirow{2}{*}{Method}   & \multicolumn{3}{c}{Top-1 Acc (\%) @ W/A} \\ \cline{2-4} 
                                  & 2 / 2    & 3 / 3   & 4 / 4   \\ \clineB{1-4}{2}
        LSQ \cite{LSQ_Steven}     & 66.5     & 69.4    & 70.5    \\ \hline
        Ascending layer training  & 66.5     & 67.9    & 68.4    \\ \hline
        Multi-layer co-training   &          & NaN     &         \\
        \toprule[1pt]
        \end{tabular}
    }
    \end{small}
    \label{tab: baseline comp}
\end{table}

\section{Our Method}
\label{sec: our method}

\begin{figure*}[t]
	\centering
	\includegraphics[width=1\linewidth]{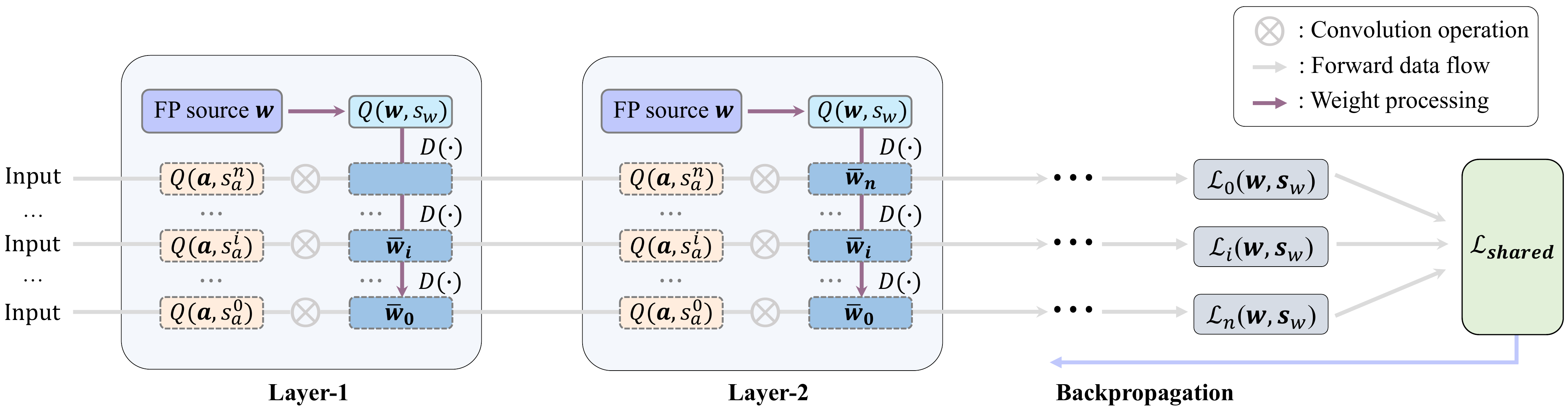}
	\vspace{-0.4cm}
        \caption{Structure of the proposed training framework. The loss set $ \{\mathcal{L}_n,\cdots,\mathcal{L}_0\}$ is calculated by feeding the training data into networks with different bit-width parameters $\{\bar{\bm{w}}_n, \cdots, \bar{\bm{w}}_0\}$. These losses are combined and gradients are backpropagated to update the source parameters and the source step size.}
        \label{fig: training framework}
\end{figure*}

This section provides the methodology for enabling high-performance vertical-layered weight representation within a single model. Our method includes the downsampling quantization scheme (Sec.~\ref{subsec: downsampling quantization}) to allow vertical-layered representation and the joint training paradigm (Sec.~\ref{subsec: joint training approach}) to optimize multiple networks simultaneously.

\subsection{Downsampling Quantization}
\label{subsec: downsampling quantization}

As mentioned in Sec.~\ref{subsec: naive approaches}, it is challenging to directly learn all discrete vertical layers since the difficulty in optimizing multiple sets of discrete parameters simultaneously. 
Instead, we build a direct mapping between two adjacent precision quantized weights avoiding the direct optimization of separate enhance layers.
The detailed illustration is shown in Fig.~\ref{fig: training framework}. 
First, we quantize the full precision weights $\bm{w}$ into the highest desired bit width $\bar{\bm{w}}_n$ through Eq.~\eqref{eq: uniform quantization}. 
We denote the step size $s_w$ for quantization and set it as the step size of the highest precision quantized network, \ie $s_n = s_w$.
Then, we introduce the recursive downsampling operation $D(\cdot)$ to generate the lower bit-width quantized weights $\bar{\bm{w}}_i$, 
\begin{equation}
    \bar{\bm{w}}_{i} = D(\bar{\bm{w}}_{i+1}) = \lfloor \frac{\bar{\bm{w}}_{i+1}}{2} \rfloor, \quad i \in \{n-1, \cdots, 0 \};
\end{equation}
where we use the floor function $\lfloor \cdot \rfloor$ instead of the rounding function $\lfloor \cdot \rceil$ for obtaining the lower bit-width quantized value to satisfy the requirements of vertical-layered representation and avoid the data overflow 
\footnote{The rounding operation may not be able to reduce the bit width and thus violates the principles of the vertical-layered model. For example, given a 3-bit value $\bar{w}_i = 7$, the floor quantized value $\lfloor \frac{\bar{w}_{i}}{2} \rfloor= 3$ is in 2-bit, while the rounding quantized value $\lfloor \frac{\bar{w}_{i}}{2} \rceil = 4$ is still in 3-bit.}. 
Thus, the $i$-th binary enhance layer could be obtained by $\bm{b}_i = \bar{\bm{w}}_i - 2 \times \bar{\bm{w}}_{i-1}$. 
The downsampling operation satisfies the requirement of a vertical-layered representation since the mapping between $\bar{\bm{w}}_{i}$ and $\bar{\bm{w}}_{i-1}$ is exactly based on finding and discarding the least significant bit $\bm{b}_i$ of $\bar{\bm{w}}_i$. 
The downsampling operation can be applied recursively until the quantized weight achieves the lowest bit width, \ie the basic layer $\bar{\bm{w}}_0$. 
Therefore, the quantized weight integrated by $i$ enhance layers can be obtained by applying downsampling on $\bar{\bm{w}}_n$ with $n-i$ times, which is further used to compose a neural network running at the specific precision as illustrated in Fig.~\ref{fig: scalable inference}. 
The enhance layers $\{\bm{b}_n, \bm{b}_{n-1}, \cdots, \bm{b}_1\}$ can be obtained by collecting the dropped bits along the sequential downsampling process.  
However, the direct employment of downsampling operation for obtaining the desired bit-width quantized weight will significantly deteriorate performance, see Fig.~\ref{fig: weight histogram comparison}. We propose the following positive compensation scheme to address this issue.

\vspace{0.2cm}
\noindent \textbf{Positive compensation.} 
By comparing the low bit-width parameter distribution histogram, we find two main issues: \textit{1).} most quantized weights become inactivated (\ie zero); \textit{2).} the portion of parameters with positive value has been significantly reduced.  
The reason stems from the biased weight quantization introduced by direct downsampling. 
Specifically, the distributions of the parameters through traditional training can be approximated well by a zero-mean Bell-shaped distribution \cite{PPLQ_Fang}; and the quantized weight through dedicated QAT mimics such distribution, see Fig.~\ref{fig: weight histogram comparison}(a). 
While in Fig.~\ref{fig: weight histogram comparison}(b), the direct use of downsampling quantization causes significant disturbance to the weight distribution: it makes most weights inactivate by grouping a large portion of weights and mapping them to zero value. This thus aggravates the asymmetry of parameter distributions, \ie the generated positive weights account for only a tiny proportion compared with the one from the oracle results (Fig.~\ref{fig: weight histogram comparison}(a)).
The above problems become more severe as the bit width decreases, especially for the 2-bit case. 
Consequently, the model's capability is harmed, making the performance significantly degraded.

The above analysis motivates us to revisit a common practice in biased quantization of neural networks \cite{bias_correction, fight_bias} to compensate for the unbalanced weight distribution.
Specifically, for each bit-width quantized weight $\bm{w}_i$, we introduce a preset offset parameter for conducting the convolution operation. As illustrated in Fig.~\ref{fig: scalable inference}, the process is expressed as: 

\begin{equation}
\begin{aligned}
    \bm{y} &= (\bar{\bm{w}}_i + z_i ) \otimes \bar{\bm{a}} \times s_i \times s_a; \\
           &= (\underbrace{\bar{\bm{w}}_i \otimes \bar{\bm{a}}}_{\text{ convolution}} + \underbrace{\tilde{z}_i \times \bar{\bm{a}}}_{\text{overhead}} ) \times s_i \times s_a;
\end{aligned}
\label{eq: shift downsampling}
\end{equation}

where $\tilde{z}_i$ is the effective value to make the scalar-tensor convolution operation yielding the same result through multiplication.
In the vertical-layered representations, the compensation alleviates the weight asymmetry after downsampling, which can be seen as shifting the de-quantized weight in the positive direction a certain distance to match the original weight distribution; see Fig.~\ref{fig: weight histogram comparison}(c). Besides, this process will not incur significant computation overheads as the additional term is a low-cost scalar-tensor multiplication, which can be further treated as a bias term.

\subsection{Model Training}
\label{subsec: joint training approach}

Equipped with the downsampling scheme for the vertical-layered representation, we only need to learn the full precision weights ${\bm{w}}$ and the source step size $s_w$ to meet deployment requirements of different bit widths in the vertical-layered mode.
In this subsection, we design an effective joint training strategy with a self-knowledge distillation paradigm to learn a high-performance vertical-layered model.

\subsubsection{Joint Training Framework}
\label{subsec: training framework}

To alleviate the conflicts among quantized networks with different bit widths, the joint training process uses the loss functions of all bit-width networks to guide the learning process. 
The overall training framework is illustrated in Fig.~\ref{fig: training framework}, where $Q(\cdot, \cdot)$ denotes the quantization operation using Eq.~\eqref{eq: uniform quantization}, and $D(\cdot)$ represents the proposed downsampling operation as described in Sec.~\ref{subsec: downsampling quantization}.
In the forward path, we calculate the cross-entropy loses $\{\mathcal{L}_n, \cdots, \mathcal{L}_0\}$ by feeding the training data into networks and producing the outputs using parameters of different bit widths $\{\bar{\bm{w}}_n, \cdots, \bar{\bm{w}}_0\}$. 
These losses would be combined to a global loss $\mathcal{L}_\text{shared}$, which will be discussed in the following Sec.~\ref{subsubsec: moo}. 
In the backward pass, we use the straight-through estimator \cite{STE_Krizhevsky} to approximate the gradient of the non-differentiable operation, including the quantization operation $Q(\cdot,\cdot)$ and downsampling operation $D(\cdot)$. 
Specifically, for a specific full precision source weight $w$, the gradient of $\mathcal{L}_i$ with respect to it is approximated via 
\begin{equation}
    \frac{\partial \mathcal{L}_i}{\partial w} \approx \frac{\partial \mathcal{L}_i}{\partial \bar{w}_i} \times \frac{1}{s_i}.
\label{eq:weight_gradient}
\end{equation}
For the corresponding step size $\{s_0, s_1, ..., s_n\}$, we can calculate the gradient of $\mathcal{L}_i$ with respect to $s_i$ through $\bar{w}_i$ as,
\begin{equation}
    \frac{\partial \mathcal{L}_i}{\partial s_i}\bigg \vert_{\bar{w}_i} \approx
    \begin{cases} 
    \begin{aligned}
    & \frac{\partial \mathcal{L}_i}{\partial \bar{w}_i} \times \frac{1}{s_i} \times (-w/s_i + \bar{w}_i), & -Q_N < \bar{w}_i < Q_P; \\
    & \frac{\partial \mathcal{L}_i}{\partial \bar{w}_i} \times \frac{1}{s_i} \times(-Q_n^N), &  \bar{w}_i < -Q_N;  \\
    & \frac{\partial \mathcal{L}_i}{\partial \bar{w}_i} \times \frac{1}{s_i} \times(Q_p^N) , & \bar{w}_i > Q_P.  \\
    \end{aligned}
    \end{cases}
    \label{eq:step_size_gradient}
\end{equation}
Since the step sizes are coupled by the source quantization step size $s_w$,
the gradient of $\mathcal{L}_i$ with respect to the source parameter $s_w$ through specific $\bar{w}_i$ can be successively backpropagated for optimization.
Then, the gradients from different losses are combined and further used to update source model weights $\bm{w}$ and the step size $s_w$. In the following section, we will elaborate on combining losses/gradients from different losses into a multi-objective optimization (MOO) framework to achieve the balance among optimizing multiple objectives.

\subsubsection{Multi-Objective Optimization}
\label{subsubsec: moo}

It is critical to find a good trade-off between different objectives $\{\mathcal{L}_0, \mathcal{L}_1, \cdots, \mathcal{L}_n\}$ to ensure all intermediate quantized models could achieve competitive performance. 
To achieve this, we define the following multi-objective optimization (MOO) problem,
\begin{equation}
    \mathcal{L}_\text{shared} = \sum_{i=0}^n \alpha_i \mathcal{L}_i(\bm{w}, \bm{s}_w);
    \label{eq: unitary scalarization}
\end{equation}
where $\alpha_i$ is the scale factor of the corresponding loss function $\mathcal{L}_i(\bm{w}, \bm{s}_w)$, $\bm{w}$ and $\bm{s}_w$ denotes the set of all full precision weights and the source step size. 
We investigate two well-known baseline optimizers for the MOO problem: the unitary scalarization \cite{US_Kurin} and the multiple gradient descent \cite{MGDA_Desideri}.

\vspace{0.2cm}
\noindent\textbf{Unitary scalarization.} Unitary scalarization (US) \cite{US_Kurin} is a recently proposed multi-task learning framework, which suggests that using the summation of losses with appropriate scale factors is sufficient to obtain comparable results as the specialized multi-objective optimizers. 
It is suggested that the interference among multiple losses can be treated as a form of regularization \cite{US_Kurin}. Without introducing any priors on different objectives, we empirically set $\alpha_i = \sfrac{1}{(n+1)}$ to calculate the combined loss in the following experiments, see Sec.~\ref{sec: experiments}.

\vspace{0.2cm}
\noindent \textbf{Multiple gradient descent.} The multiple gradient descent (MGD) algorithm demonstrates that a smaller norm of the convex combination of gradients indicates minor interferences among different objectives by analyzing the KKT conditions \cite{MGDA_Desideri}. It thus dynamically updates the combination factors $\alpha_i$ by minimizing the overall gradient norm as:
\begin{equation}
\begin{aligned}
    \mathop{\arg \min}_{\alpha_0, \cdots, \alpha_n} \bigg\Vert \sum_{i=0}^{n} \alpha_i \frac{\partial \mathcal{L}_i(\boldsymbol{w}, \boldsymbol{s}_w)}{ \partial (\boldsymbol{w}, \boldsymbol{s}_w)} \bigg\Vert_2^2,
    \quad \text{s.t.} ~\alpha_i \geq 0 ~\text{and} \sum_{i=0}^n \alpha_i = 1; 
\label{eq: mgd}
\end{aligned}
\end{equation}
where the gradient of $\boldsymbol{w}$ and $\boldsymbol{s_w}$ are calculated via Eq.~\eqref{eq:weight_gradient} and Eq.~\eqref{eq:step_size_gradient} respectively.
The work in \cite{MTL_MGDA} provides an effective update rule for $\alpha_i$ based on the Frank-Wolfe algorithm.

Besides the shared parameters above, our framework has some bit-width-specific parameters, including the batch normalization parameters (\ie mean, variance, and affine parameters) and all activation quantization step sizes.
The mean and variance are updated in a moving average manner for each bit width separately. 
The affine parameters and step sizes for activations are optimized using the loss of the corresponding bit-width network loss $\mathcal{L}_i$, which we do not include in the MOO framework.

\subsubsection{Self-Knowledge Distillation}
\label{subsubsec: self-knowledge distillation}

It is a broad consensus that knowledge distillation is beneficial to improve the quantized models' performance \cite{QKD_Kim, QAT_NAS_Shen}. 
Here, we adopt the self-knowledge distillation to further improve model performance, where the outputs from a high bit-width network (teacher) are directly used to supervise the lower bit-width network (student). The overall loss function is formulated as
\begin{equation}
    \mathcal{L}_i (\boldsymbol{w}, \boldsymbol{s}_w) = \mathcal{L}_{\text{CE}}^i + \mathcal{L}_{\text{KD}}(\mathbf{z}_{i+1}, \mathbf{z}_i); \quad i < n.
    \label{eq: kd}
\end{equation}
$\mathcal{L}_{\text{CE}}^i$ is the original cross-entropy training loss, $\mathcal{L}_{\text{KD}}$ is the knowledge distillation function, $\mathbf{z}_{i+1}$ and $\mathbf{z}_{i}$ denote the probabilistic softmax output from the teacher and the student respectively.
We apply the cosine similarity function for constructing the distillation $\mathcal{L}_{\text{KD}}$, which has been demonstrated to be effective when the teacher's prediction is ambiguous \cite{SKDQ_Boo}.

\section{Discussion}
\label{Sec: discussion}

\noindent 
\textbf{Mixed-precision quantization.} The mixed-precision quantization network allows different layers residing in one network to adopt different bit-width computations for flexibility, especially when there are constraints for a model size or the total number of operations \cite{HAQ_Wang}.
As our vertical-layered model is trained in a uniform-precision manner, we wonder if extending the obtained model in a mixed-precision setting is possible. Since the quantization error is a commonly adopted criterion for optimizing the bits allocation among layers within one network, we apply a similar criterion that minimizes the overall downsampling error constrained on a total number of bits budget, \ie
\begin{equation}
\begin{aligned}
     \mathop{\arg \min}_{{i_m} \in \{0, \cdots, n\}} & \sum_{m=1}^{M} \Vert \bar{\bm{w}}_{i_m} - \bar{\bm{w}}_{n_m} \Vert_2^2, & \\
     \text{s.t.} \qquad & \sum_{m=1}^{M} N_m (|\bar{\bm{w}}_{i_m}| - |\bar{\bm{w}}_{n_m}|) \leq B; &
     \label{eq: mixed precision sampling}
\end{aligned}
\end{equation} 
where $\bar{\bm{w}}_{i_m}$ is the quantized weight tensor integrated with $i$ enhance layers in the network's $m$-th layer, $|\bar{\bm{w}}_{i_m}|$ is the bit width assigned to every entry in the $m$-th layer, $N_m$ is the corresponding number of weight parameters and $B$ is the total bits budget. The detailed implementation and analysis will be covered in Sec.~\ref{sec: mixed precision}.

\vspace{0.2cm}
\noindent 
\textbf{Comparison with NAS-based approaches.} 
Neural architecture search \cite{NAS_SP_Guo, HAQ_Wang, HAWQ_Dong, HAWQ_V3_Yao, APQ_Wang, QAT_NAS_Shen} is a widely adopted technique in the mixed-precision quantization, which is orthogonal to our effort. For instance, by incorporating NAS to search for a more suitable bit width assignment strategy, the obtained model can attain better performance, which is not the focus of this paper. 
Besides, the model yielded from pure NAS-based approaches cannot offer graceful on-demand model upgrade/downgrade without accessing the superfluous source parameters.
For instance, the NAS-based quantization method represented by \cite{QAT_NAS_Shen} requires considerable computing resources and time to search the network structure for the specified bit-width, and the searched model also needs further careful fine-tuning to achieve satisfactory performance. In contrast, our method does not require a complex fine-tuning process and heavy searching process.

\section{Experiments}
\label{sec: experiments}

\vspace{0.2cm}
\subsection{Experimental Setup}
\label{sec:exp_setup}
We conduct extensive experiments on CIFAR-100 \cite{CIFAR_Krizhevsky} and ImageNet ILSVRC2012 \cite{ImageNet_Russakovsky}. Experimental setups are shown as follows. 

\vspace{0.2cm}
\noindent \textbf{Comparison setup.} 
Unlike our work, which mainly focuses on training vertical-layered networks, 
prior QAT works aim at optimizing the model for a specific bit width.
We incorporate multiple state-of-the-art QAT approaches for comparisons, including PACT \cite{PACT_Choi}, LQ-Nets \cite{LQNet_Zhang}, DSQ \cite{DSQ_Gong}, QIL \cite{QIL_Jung}, LQW \cite{LQW_Hoang}. 
Besides, we implement the quantized networks tailored to specific bit-width via QAT using LSQ \cite{LSQ_Steven}, which serves as the oracle results and sets an upper bound performance of our model.

\begin{table}[t]
\renewcommand\arraystretch{1.4}
    \centering
    \caption{Results of different quantization approaches on CIFAR-100 are presented. The network architecture used here is the ResNet-18. The network's weights and the corresponding layers' input are quantized into W/A-bit separately. The results marked with $^*$ are from \cite{DiffQ}.}
    \begin{small}
    \setlength{\tabcolsep}{3.9mm}{
        \begin{tabular}{l|ccc|c}
        \bottomrule[1pt]
        \multicolumn{1}{c|}{\multirow{2}{*}{\begin{tabular}[c]{@{}c@{}}Quantization \\ scheme\end{tabular}}} & \multicolumn{3}{c|}{Top-1 Acc (\%) @ W/A} & \multirow{2}{*}{\begin{tabular}[c]{@{}c@{}}Vertical- \\ layered\end{tabular}} \\ \cline{2-4}
        \multicolumn{1}{c|}{}   & 2 / 2  & 3 / 3  & 4 / 4    & \\
        \clineB{1-5}{2.5}
        LQ-Nets$^{*}$ \cite{LQNet_Zhang}   & 70.8           & 72.9  & -     & \xmark        \\ \hline
        LQW$^{*}$ \cite{LQW_Hoang}         & 65.1           & 73.0  & -     & \xmark        \\ \hline
        DiffQ$^{*}$ \cite{DiffQ}           & 66.6           & 76.7  & 77.5  & \xmark        \\ \hline
        LSQ \cite{LSQ_Steven}              & \textbf{76.9}  & 77.7  & 77.8  & \xmark        \\ \hline 
        Ours                               & 76.3           & \textbf{77.8} & \textbf{78.3}  & \cmark  \\
        \toprule[1pt]
        \end{tabular}
        }
    \end{small}
    \label{tab: cifar100 comp}
\end{table}

\begin{table*}[t]
\renewcommand\arraystretch{1.4}
    \centering
    \caption{Comparison of different quantization approaches on ImageNet dataset. Top-1 and Top-5 accuracy (\%) are reported separately. The network's weight and the corresponding layers' input are quantized into W/A-bit separately. The full precision (FP) models' Top-1 and Top-5 accuracy are attached in each model. Results marked with $^\dagger$ are from the original paper. Results marked with $ ^\S$ are from \cite{MQBench}.}
    \begin{small}
    \setlength{\tabcolsep}{4.9mm}{
        \begin{tabular}{c|l|ccc|ccc|c}
            \bottomrule[1pt]
            \multirow{2}{*}{\begin{tabular}[c]{@{}c@{}}Network \\ model\end{tabular}} & \multicolumn{1}{c|}{\multirow{2}{*}{\begin{tabular}[c]{@{}c@{}} Quantization \\ scheme\end{tabular}}} & \multicolumn{3}{c|}{Top-1 Acc (\%) @ W/A} & \multicolumn{3}{c|}{Top-5 Acc (\%) @ W/A} & \multirow{2}{*}{\begin{tabular}[c]{@{}c@{}} Vertical- \\  layered \end{tabular}} \\ \cline{3-8}
            & \multicolumn{1}{c|}{} & 2 / 2  & 3 / 3  & 4 / 4  & 2 / 2  &  3 / 3 & 4 / 4        \\                                                
            \clineB{1-9}{2.5}
            \multirow{7}{*}{\begin{tabular}[c]{@{}c@{}} ResNet-18 \\ \textit{FP: 70.5 / 89.6} \end{tabular}}                                               
            & PACT$^{}\dagger$ \cite{PACT_Choi}        &64.4  &68.1  &69.2  &85.6  &88.2  &89.0 &\xmark \\ \cline{2-9} 
            & LQ-Nets$^{\dagger}$ \cite{LQNet_Zhang}   &64.9  &68.2  &69.3  &85.9  &87.9  &88.8 &\xmark \\ \cline{2-9} 
            & DSQ$^{\dagger}$ \cite{DSQ_Gong}          &65.2  &68.7  &69.6  &-     &-     &-    &\xmark \\ \cline{2-9} 
            & QIL$^{\dagger}$ \cite{QIL_Jung}          &65.7  &69.2  &70.4  &-     &-     &-    &\xmark \\ \cline{2-9} 
            & LSQ \cite{LSQ_Steven}        &66.5  &69.4  &70.5  &87.0  &88.9  &\textbf{89.5} &\xmark \\ \cline{2-9} 
            & Ours (\textit{w/o} self-KD)           &66.7  &69.9  &\textbf{70.6}  &\textbf{87.3}  &\textbf{89.1}  &\textbf{89.5} &\cmark \\ \cline{2-9}
            & Ours (\textit{w.} self-KD)            &\textbf{67.1}  &\textbf{70.1}  &\textbf{70.6}  &87.2  &\textbf{89.1}  &\textbf{89.5} &\cmark \\
            \clineB{1-9}{2.5}
            \multirow{6}{*}{\begin{tabular}[c]{@{}c@{}} ResNet-34 \\  \textit{FP: 74.1 / 92.8} \end{tabular}}                                                
            & LQ-Nets$^{\dagger}$ \cite{LQNet_Zhang}  &69.8  &71.9  &-     &89.1  &90.2  &-    &\xmark \\ \cline{2-9} 
            & DSQ$^{\dagger}$ \cite{DSQ_Gong}         &70.0  &72.5  &72.8  &-     &-     &-    &\xmark \\ \cline{2-9} 
            & QIL$^{\dagger}$ \cite{QIL_Jung}         &70.6  &73.1  &73.7  &-     &-     &-    &\xmark \\ \cline{2-9} 
            & LSQ \cite{LSQ_Steven}       &71.1  &\textbf{73.6}  &\textbf{74.1}  &\textbf{90.0}  &\textbf{91.5}  &\textbf{91.6} &\xmark \\ \cline{2-9}
            & Ours (\textit{w/o} self-KD)          &70.7  &73.3  &73.9  &89.8  &91.3  &\textbf{91.6} &\cmark \\ \cline{2-9} 
            & Ours (\textit{w.} self-KD)           &\textbf{71.2}  &73.5  &74.0  &89.8  &91.3  &91.5 &\cmark \\
            \clineB{1-9}{2.5}
            \multirow{5}{*}{\begin{tabular}[c]{@{}c@{}} ResNet-50 \\  \textit{FP: 76.9 / 93.4} \end{tabular}}  
            & PACT$^{\dagger}$ \cite{PACT_Choi}       &72.2  &75.3  &76.5  &-     &-     &-    &\xmark \\ \cline{2-9} 
            & LQ-Nets$^{\dagger}$ \cite{LQNet_Zhang}  &71.5  &74.2  &75.1  &90.3  &91.6  &92.4 &\xmark \\ \cline{2-9} 
            & LSQ \cite{LSQ_Steven}       &73.2  &75.8  &\textbf{76.8}  &\textbf{91.3}  &92.6  &93.0 &\xmark \\ \cline{2-9} 
            & Ours (\textit{w/o} self-KD)          &72.4  &75.9  &\textbf{76.8}  &90.9  &92.7  &93.0 &\cmark \\ \cline{2-9}
            & Ours (\textit{w.} self-KD)           &\textbf{73.2}  &\textbf{76.1}  &\textbf{76.8}  &91.1  &\textbf{92.8}  &\textbf{93.1} &\cmark \\
            \clineB{1-9}{2.5}
            \multirow{4}{*}{\begin{tabular}[c]{@{}c@{}} MobileNetV2 \\  \textit{FP: 71.9 / 91.3} \end{tabular}}  
            & PACT$^{\dagger}$ \cite{PACT_Choi}       &-  &-     &61.4     &-  &-     &83.7    &\xmark \\ \cline{2-9} 
            & LSQ$^{\S}$ \cite{LSQ_Steven}         &46.7  &-   &66.3  &-  &-  &- &\xmark \\ \cline{2-9} 
            & Ours (\textit{w/o} self-KD)          &47.4  & 64.3   & 67.1  & 72.5  & 85.5   &\textbf{87.8} &\cmark \\ \cline{2-9}
            & Ours (\textit{w.} self-KD)           &\textbf{50.6}  &\textbf{65.0}  & \textbf{67.5} &\textbf{75.3}  &\textbf{86.0} & \textbf{87.8} &\cmark \\
            \clineB{1-9}{2.5}
            \multirow{3}{*}{\begin{tabular}[c]{@{}c@{}} VGG-16 BN \\  \textit{FP: 73.4 / 91.5} \end{tabular}}  
            & LSQ$^{\dagger}$ \cite{LSQ_Steven}    &\textbf{71.4}  &\textbf{73.4}   & \textbf{74.0}  &\textbf{90.4}  &\textbf{91.5}  &\textbf{92.0}  &\xmark \\ \cline{2-9} 
            & Ours (\textit{w/o} self-KD)          &70.9  &73.0   &73.4  &90.1  &91.3  &91.6  &\cmark \\ \cline{2-9}
            & Ours (\textit{w.} self-KD)           &71.0  &73.3  &73.4  &89.9  &91.2  &91.4 &\cmark \\
            \toprule[1pt]
            \end{tabular}
        }
    \end{small}
    \label{tab: imagenet comp}
\end{table*}

\vspace{0.2cm}
\noindent \textbf{Training framework.} 
The overall vertical-layered training framework is shown in Fig.~\ref{fig: training framework}. 
The full precision weights $\bm{w}$ and the source step sizes $\bm{s}_w$ are learned under the MOO framework introduced in Sec.~\ref{subsubsec: moo}. 
Different bit-width networks hold separate sets of quantization step sizes for the quantization of activations, which are learned through the specific loss $\mathcal{L}_i$ as detailed in \cite{LSQ_Steven}. 
By default, the batch normalization layers of networks with different bit widths are separate. 
Following the common practice \cite{XOR_Rast, QNN_Itay, PACT_Choi, LSQ_Steven}, the first and last layers of a DNN model are not quantized, and the weight parameters of these two layers are shared among different bit-width networks. 
We set the basic layer of our model as a 2-bit quantized network instead of 1-bit because the binary neural networks usually require dedicated network structures, which are incompatible with higher bit width quantization structures \cite{BNN_Cour, XOR_Rast}. 
The highest bit-width in our model is set to be 4-bit since a 4-bit quantized network can achieve a full precision network's performance \cite{PACT_Choi}.

\vspace{0.2cm}
\noindent \textbf{Image pre-processing.} 
For the CIFAR dataset, we follow a standard procedure in \cite{WResNet_Sergey} where training images are zero-padded with 4 pixels on every side horizontally and vertically, and then a random $32 \times 32$ patch is cropped, which is followed by a random horizontal flip with a probability of $0.5$. 
For the ImageNet dataset, images in the training set are randomly resized and cropped to $224 \times 224$ with a random horizontal flipping probability of $0.5$. The images in the validation set are resized to $256 \times 256$ and center copped to $224 \times 224$ for evaluation.

\vspace{0.2cm}
\noindent \textbf{Hyper-parameters.} 
All networks are trained using stochastic gradient descent with a momentum of $0.9$ using the cosine learning rate decay schedule without restart \cite{SGDR_Loshchilov}. 
The softmax cross-entropy loss is applied to guide the learning process. 
We first train the full precision networks from scratch and then conduct the quantization-aware training. 
For the CIFAR dataset, we train 300 epochs with an initial learning rate of $0.1$, batch size $128$, and weight decay $1e-3$; the initial learning rate is scaled by $0.1$ for training quantized networks.
For ImageNet, the initial learning rate is $0.1$, and batch size is $256$ for ResNet-18 and ResNet-34; for ResNet-50, the initial learning rate is $0.05$ with batch size $128$. 
The weight decay is $1e-4$ for full precision network training. 
For quantized networks, all initial learning rates are set to be $0.01$, and the same weight decay and batch size as full precision training are applied, except for a weight decay of $5e-5$ in ResNet-18.
In all experimented settings, we simply use a preset compensation value as $z_i = 2^{-1} (1 - 2 ^ {i-2})$ to minimize the downsampling error, which is demonstrated to be sufficient for obtaining comparable accuracy as the bit-width tailored QAT results.
The oracle bit-width tailored quantized neural networks are trained with the same fixed gradient scaling factors for fair comparisons.

\vspace{0.2cm}
\noindent \textbf{Initialization.} We initialize the network weights and all batch normalization parameters using well-trained full precision networks. The full precision MobileNetV2 \cite{MobilenetV2_Sandler} and VGG-16 \cite{VGG} are from the Pytorch library \cite{Pytorch_Paszke}.
For the quantization step size of weights, we initialize them using averaged absolute values of full precision weights, \ie $s_w = \sfrac{2 \times mean\{|\bm{w}|\}}{\sqrt{2^{4} - 1}}$.
For the quantization step size of activation, we use the first iteration's input activation to initialize every step size by $s_a^i = \sfrac{\max\{|\bm{a}|\}}{(2^{i+1} - 1)} $.

\subsection{Main results and Analysis}
\label{main_results}

\vspace{0.1cm}
\subsubsection{Experiments on CIFAR-100}

We experiment with ResNet-18 \cite{ResNet_He} using different quantization methods on the CIFAR-100 dataset. The Top-1 validation accuracy is shown in Table~\ref{tab: cifar100 comp}. 
The network weights and the corresponding input are quantized into W-bit and A-bit, respectively, denoted as W/A. 
Our method significantly surpasses many baselines. 
For example, it exceeds LQ-Net \cite{LQNet_Zhang} by 5.5\% under the 2-bit setting, and outperforms DiffQ \cite{DiffQ} by 1.1\% and 0.8\% under the 3-bit and 4-bit setting respectively. 
Moreover, our method shows comparable performance even in comparison with the bit-width tailored QAT using LSQ\cite{LSQ_Steven}. 
Compared with all baselines, only our method has the vertical-layered characteristic, that is, once training and scalable inference for various bit-width, while does not require separate quantization and QAT training tailored to specific bit-width as the baseline does.
We can conclude that: \textit{1).} our vertical-layered representation enables one-time training for scalable inference; \textit{2).} our proposed joint training design can support vertical-layered settings well without sacrificing model performance.

\vspace{0.1cm}
\subsubsection{Experiments on ImageNet}

\begin{table*}[t]
\renewcommand\arraystretch{1.4}
    \centering
    \caption{Results of using different batch normalization layers on ImageNet are compared. We summarize the number (M) of BN parameters under different BN settings and list their proportion (\textperthousand) in the total number of model parameters. Top-1 and Top-5 accuracy (\%) are reported separately.}
    \begin{small}
    \setlength{\tabcolsep}{4.2mm}{
    \begin{tabular}{c|l| c |ccc c ccc}
    \bottomrule[1pt]
    \multirow{2}{*}{Model} & \multirow{2}{*}{BN Setting} & \multirow{2}{*}{\begin{tabular}[c]{@{}c@{}} \# of BN \\ Parameters \end{tabular}} & \multicolumn{3}{c}{Top-1 Acc (\%) @ W / A} & & \multicolumn{3}{c}{Top-5 Acc (\%) @ W / A} \\ \cline{4-10} 
    & & & 2 / 2 & 3 / 3 & 4 / 4  &  & 2 / 2 & 3 / 3 & 4 / 4 \\ 
    \clineB{1-10}{2.5}
    \multirow{3}{*}{ResNet-18} 
    & Shared BN               & 0.19M (1.6\textperthousand) & 33.5 & 67.4 & 49.3 &  & 58.0 & 87.8 & 74.1 \\
    & + Individual Statistics & 0.38M (3.3\textperthousand) & 66.6 & 68.8 & 69.2 &  & 86.9 & 88.6 & 88.8 \\
    & ++ Individual Weights   & 0.58M (4.9\textperthousand) & \textbf{66.7} & \textbf{69.9} & \textbf{70.6} &  & \textbf{87.3} & \textbf{89.1} & \textbf{89.5} \\
    \clineB{1-10}{2.5}
    \multirow{3}{*}{ResNet-34} 
    & Shared BN               & 0.34M (1.5\textperthousand) & 56.2 & 72.2 & 41.1 &  & 79.7 & 90.8 & 66.6 \\
    & + Individual Statistics & 0.68M (3.0\textperthousand) & 70.7 & 72.7 & 73.0 &  & 89.8 & 91.0 & 91.3 \\
    & ++ Individual Weights   & 1.02M (4.5\textperthousand) & \textbf{70.7} & \textbf{73.3} & \textbf{73.9} &  & \textbf{89.8} & \textbf{91.3} & \textbf{91.6} \\
    \toprule[1pt]
    \end{tabular}
    }
    \end{small}
    \label{table: switchable BN}
\end{table*}

\vspace{0.1cm}
\noindent \textbf{Main results}. For the large-scale ImageNet dataset, we perform experiments on the ResNet family \cite{ResNet_He} and the MobileNetV2 \cite{MobilenetV2_Sandler}.
Table~\ref{tab: imagenet comp} shows the Top-1 and Top-5 validation accuracy with state-of-the-art quantization methods.
Among all compared methods, ours is the only one that can achieve once training for scalable inference without tailored QAT. 
The joint training framework (Sec.~\ref{sec: our method}) achieves satisfactory performance in almost all evaluated bit-width settings, and self-KD further improves the model's Top-1 accuracy by $0.1\%-0.5\%$. 
Our method shows comparable or even better performance in comparison with the bit-width-specific oracle LSQ method, especially for the lightweight model cases (ResNet-18 and MobileNetV2). For example, our joint training outperforms the oracle LSQ \cite{LSQ_Steven} by $0.6\%/0.9\%$ for 2-bit/3-bit ResNet-18 and $0.4\%/0.9\%$ for 2-bit/4-bit MobileNetV2 in terms of the Top-1 accuracy. 
In the case of a large model (ResNet-50), compared with the compared baselines, our method can effectively narrow the performance gap between 2-bit/3-bit and full precision networks, and even achieves 4-bit inference with only $0.1\%$ performance sacrifice compared with the full precision one. 

\begin{table}[h]
\renewcommand\arraystretch{1.4}
    \centering
    \caption{Results of different scalable bit-width inference approaches on ImageNet are compared.
    The network architecture used here is the ResNet-50. Networks' weight and activation are quantized into W/A-bit respectively. Results marked with $^\dagger$ are from the original paper.} 
    \begin{small}
    \setlength{\tabcolsep}{3mm}{
    \begin{tabular}{l|c|ccc}
    \bottomrule[1pt]
    \multirow{2}{*}{Scheme} & \multirow{2}{*}{\begin{tabular}[c]{@{}c@{}}Weight size\\ (bits)\end{tabular}} & \multicolumn{3}{c}{Top-1 Acc (\%) @ W/A} \\ \cline{3-5} 
                      &        & 2 / 2      & 3 / 3      & 4 / 4     \\ \clineB{1-5}{2.5}
    Ascending layer   & 159.6M &  73.2      & 73.6       &  73.9         \\ \hline
    QAT + fine-tuning   & 159.6M &  60.1      & 73.4       &  76.8        \\ \hline
    Adabits$^\dagger$ \cite{AdaBits_Jin}  &  816.1M      & \textbf{73.2}           & 75.8           & 76.1 \\ \hline
    Ours (\textit{w/o} self-KD) & 159.6M       & 72.4   & 75.9   &  \textbf{76.8}  \\ \hline
    Ours (\textit{w.} self-KD)  & 159.6M       & \textbf{73.2}   & \textbf{76.1}   & \textbf{76.8} \\ 
    \toprule[1pt]
    \end{tabular}
    }
    \end{small}
    \label{tab: Adabits comp}
\end{table}

\vspace{0.1cm}
\noindent \textbf{One model for scalable bit-width inference.} 
In this part, we compare different methods for obtaining a model that can support scalable bit-width inference. 
We implement the ascending layer approach described in Sec.~\ref{subsec: naive approaches}, fine-tuning based on the highest bit-width network obtained via QAT, and our training framework.
Here, we also compare with the related work called Adabits \cite{AdaBits_Jin}, which can be deployed for scalable bit-width inference based on a shared full precision model and is also optimized in a collaborative training manner. 
All experiments are conducted with ResNet-50 \cite{ResNet_He} on ImageNet \cite{ImageNet_Russakovsky}, and results are reported in Table~\ref{tab: Adabits comp}.
Compared with naive approaches (ascending layer and QAT + fine-tuning), our method and Adabits significantly improve the performance of all assessed bit widths, which suggests that joint training is beneficial for scalable bit-width deployment. 
With only one-fifth of the weight parameters required by Adabit, our method achieves the highest accuracy in 3-bit and 4-bit settings.
However, Adabits requires accessing the full precision parameters or storing all bit-width quantized models for scalable inference, which incurs large storage space requirements.  

\subsection{Ablation Study}

\begin{table*}[t]
\renewcommand\arraystretch{1.4}
    \centering
    \caption{Results of different MOO optimizers used in the proposed training framework are compared. Top-1 and Top-5 validation accuracy (\%) on ImageNet are reported separately. W/A indicates that the well-trained model is running on specific precision.}
    \begin{small}
    \setlength{\tabcolsep}{4.7mm}{
    \begin{tabular}{c|l| ccc c ccc}
    \bottomrule[1pt]
    \multirow{2}{*}{Model} & \multirow{2}{*}{Training approach} &  \multicolumn{3}{c}{Top-1 Acc (\%) @ W / A} & & \multicolumn{3}{c}{Top-5 Acc (\%) @ W / A} \\ \cline{3-9} 
    &       &2 / 2   &3 / 3  &4 / 4  &  &2 / 2  &3 / 3 & 4 / 4  \\ 
    \clineB{1-9}{2.5}
     \multirow{2}{*}{ResNet-18}
     & Unitary Scalarization (US) \cite{US_Kurin}         & 66.7 & \textbf{69.9} & \textbf{70.6} &  & \textbf{87.3} & \textbf{89.1} & \textbf{89.5} \\
     & Multi-Gradient Descent (MGD) \cite{MGDA_Desideri}    & \textbf{66.8} & 69.3 & 69.8 &  & \textbf{87.3} & 88.7 & 89.0 \\ 
    \clineB{1-9}{2.5}
    \multirow{2}{*}{ResNet-34} 
     & Unitary Scalarization (US) \cite{US_Kurin}          & 70.7 & \textbf{73.3} & \textbf{73.9} &  & 89.8 & \textbf{91.3} & \textbf{91.6} \\
     & Multi-Gradient Descent (MGD) \cite{MGDA_Desideri}   & \textbf{70.9} & 72.8 & 73.0 &  & \textbf{89.9} & 91.1 & 91.2 \\
    \toprule[1pt]
    \end{tabular}
    }
    \end{small}
    \label{table: ablation MOO}
\end{table*}
\vspace{0.4cm}

\begin{figure*}[t]
    \centering
    \vspace{0.2cm}
    \includegraphics[width=1\linewidth]{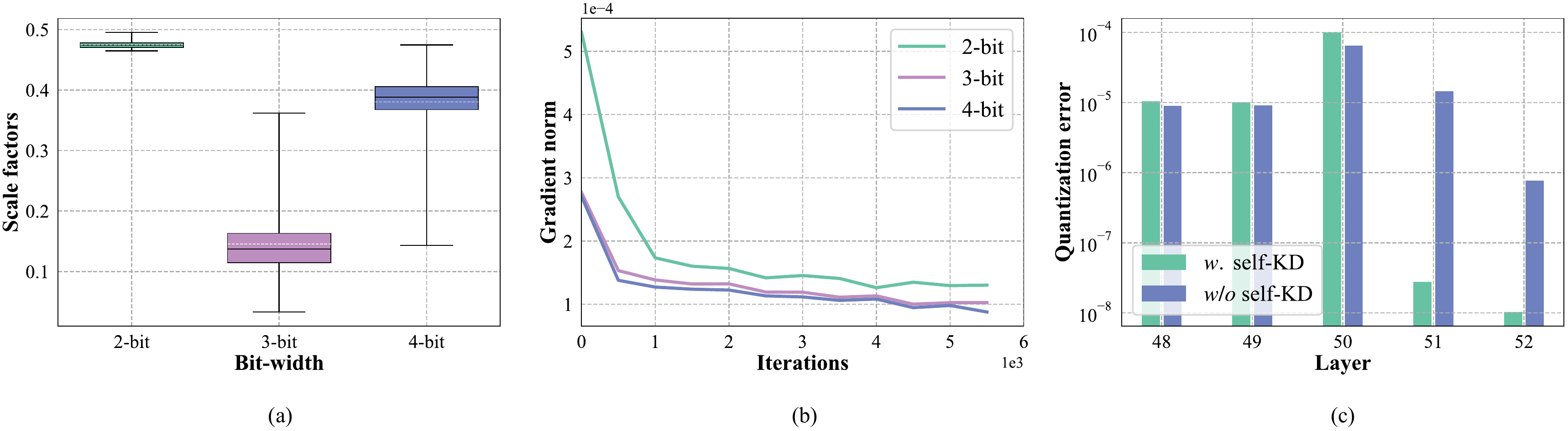}
    \vspace{-0.5cm}
    \caption{Ablation study: (a). the statistics of scale factors calculated by MGD during ResNet-18 training on the ImageNet dataset; (b). the average gradient norm during MGD-based ResNet-18 training on the ImageNet dataset; (c). the average activation quantization error norm of the last five convolution layers (layer-48 to layer-52) in 2-bit ResNet-50 models trained with/without the self-KD scheme.}
    \label{fig: ablation}
\end{figure*}

In this subsection, we conduct the following ablation experiments to assess the effectiveness of different components of our model:

\begin{itemize}
    \item Sec.~\ref{subsubsec: ablation BN} explores effective ways to construct batch normalization layers in vertical-layered models.
    \item Sec.~\ref{subsubsec: ablation positive compensation} shows the effect of the positive compensation scheme used in the downsampling operation.
    \item Sec.~\ref{subsubsec: ablation MOO} investigates different multi-objective optimizers for MOO training.
    \item Sec.~\ref{subsubsec: ablation self-KD} studies the effect of the self-knowledge distillation paradigm in MOO training.
    \item Sec.~\ref{subsubsec: ablation FP} shows the effect of using different first and last layers in vertical-layered models.
\end{itemize}

\subsubsection{Effective Ways to Construct Batch Normalization Layer}
\label{subsubsec: ablation BN}

We study three ways to construct batch normalization (BN) layers in the vertical-layered model training: 

\noindent\textit{1). Shared BN:} all batch normalization parameters, including the BN statistics (running mean and variance) and the affine parameters, are shared among networks of different bit widths.
 
\noindent\textit{2). Individual statistics:} 
individual statistics with shared BN affine parameters for different bit-width precision quantized networks are applied.

\noindent\textit{3). Individual BN:} 
BN layers are independent across different bit-width networks. This scheme has the highest flexibility and is the default setting in our main experiments. 

We experiment with ResNet-18 and ResNet-34 on ImageNet and report the comparisons in Table~\ref{table: switchable BN}. 
Experimental results show that the first choice (shared BN)  does not converge well: the 2-bit ResNet-18 only achieves 33\% top-1 accuracy, which is significantly lower than any other choice. 
Compared with the first setting, the individual statistics setting is more flexible and achieves better performance with negligible costs for increasing the number of parameters. 
The individual BN, the default setting in our experiments, achieves the highest performance among all these choices at the cost of less than 5{\textperthousand} additional parameter overhead. For example, it exceeds the first one (shared BN) by 33\% top-1 accuracy in the 2-bit ResNet-18 experiment and surpasses the second one (individual statistics) by about 1\% top-1 accuracy in the 3-bit ResNet-18 experiment.

\subsubsection{Effect of the Positive Compensation}
\label{subsubsec: ablation positive compensation}

In this part, we investigate the effect of introducing positive compensation in the downsampling process. Table~\ref{tab: ablation positive compensation} shows our experiment results with ResNet-18 and MobilNetV2 on the ImageNet dataset. We can observe that without positive compensation, the 2-bit ResNet-18 gets around a 5\% Top-1 accuracy drop compared with the compensated one, and the 3-bit network also suffers from direct downsampling. For MobileNetV2, the QAT without compensation significantly drops all the bit-width network performance, especially in the 2-bit case where the Top-1 accuracy is decreased by 10\%.
As observed from Table~\ref{tab: imagenet comp}, the joint training achieves competitive accuracy as the bit-width tailored QAT with the assistance of positive compensation in all bit-width settings.

\begin{table}[htp]
\renewcommand\arraystretch{1.4}
    \centering
    \caption{Effect of involving positive compensation in downsampling is presented. } 
    \begin{small}
    \setlength{\tabcolsep}{2.8mm}{
    \begin{tabular}{c|c|ccc}
    \bottomrule[1pt]
    \multirow{2}{*}{Model} & \multirow{2}{*}{\begin{tabular}[c]{@{}c@{}} Quantization \\ Scheme \end{tabular}} & \multicolumn{3}{c}{Top-1 Acc (\%) @ W/A} \\ \cline{3-5}&        & 2 / 2  & 3 / 3  & 4 / 4   \\ \clineB{1-5}{2.5}
    \multirow{2}{*}{ResNet-18}  
    & \textit{w/o} compensation    & 61.3   & 67.4   & 70.5    \\
    & \textit{w.} compensation     & \textbf{66.7}   & \textbf{69.9}   & \textbf{70.6}    \\
     \cline{1-5}
    \multirow{2}{*}{MobileNetV2}  
    & \textit{w/o} compensation    & 40.4   & 57.5   & 63.6   \\
    & \textit{w.} compensation     & \textbf{50.6}   & \textbf{65.0}   & \textbf{67.5}    \\
    \toprule[1pt]
    \end{tabular}
    }
    \end{small}
    \label{tab: ablation positive compensation}
\end{table}

\begin{table*}[t]
\renewcommand\arraystretch{1.4}
    \centering
    \caption{Effect of replacing full precision first convolutional and final fully-connected layer. The additional number of parameters for the first convolutional layer and final FC are summed. Top-1 and Top-5 accuracy (\%) at bit-width W/A are reported separately.}
    \begin{small}
    \setlength{\tabcolsep}{4.5mm}{
    \begin{tabular}{c|c|c|ccc c ccc}
    \bottomrule[1pt]
    \multirow{2}{*}{Model} & \multirow{2}{*}{Replaced layers} & \multirow{2}{*}{\begin{tabular}[c]{@{}c@{}} \# of additional \\ parameters \end{tabular}} & \multicolumn{3}{c}{Top-1 Acc (\%) @ W / A} & & \multicolumn{3}{c}{Top-5 Acc (\%) @ W / A} \\ \cline{4-10} 
    & & & 2 / 2 & 3 / 3 & 4 / 4  &  & 2 / 2 & 3 / 3 & 4 / 4 \\ 
    \clineB{1-10}{2.5}
    \multirow{2}{*}{ResNet-18} 
    & \xmark & - & 66.7 & 69.9 & \textbf{70.6} &  & \textbf{87.3} & 89.1 & \textbf{89.5} \\
    & \cmark & 1.04 M (9.4\%) & \textbf{66.9} & \textbf{70.0} & 70.5 &  & \textbf{87.3} & \textbf{89.3} & 89.4 \\
    \clineB{1-10}{2.5}
    \multirow{2}{*}{ResNet-50} 
    & \xmark & - & \textbf{72.4} & \textbf{75.9} & \textbf{76.8} &  & \textbf{90.9} & \textbf{92.7} & \textbf{93.0} \\
    & \cmark & 4.12 M (17.6\%)  & 71.0 & 74.8 & 75.8 &  & 90.3 & 92.3 & 92.7 \\
    \toprule[1pt]
    \end{tabular}
    }
    \end{small}
    \label{table: ablation FP}
\end{table*}

\subsubsection{Choice of the MOO Optimizer} 
\label{subsubsec: ablation MOO}

The proposed vertical-layered training framework should find a balance between the losses of different bit-width networks, which constitutes a multi-objective optimization (MOO) problem. 
We use the unitary scalarization (US) as the MOO optimizer by default. 
We further explore whether other MOO optimizers, such as the popular multi-gradient descent (MGD), can make our model achieve better results. 
The MGD method iteratively updates the combination weight $\alpha_i$ by solving the min-norm point problem detailed in Eq.~\eqref{eq: mgd}.

We present the results in Table~\ref{table: ablation MOO}. Although MGD is developed to benefit multi-task learning, it will not necessarily boost network performance compared to the simple US approach or even degrade the performance. 
In particular, MGD shades the performance significantly in 3-bit and 4-bit cases. 
We attribute this to the following two reasons:
1). MGD is not a stable choice here. Fig.~\ref{fig: ablation}(a) shows that the factors obtained by MGD have significant variances, which are unstable or even change dramatically since the inexact gradient is calculated via the unreliable biased STE \cite{STE_Krizhevsky} gradient estimator; 
2). Fig.~\ref{fig: ablation}(b) shows that the gradient norm of the 2-bit parameter is of the greatest magnitude, especially in the early training stage. However, MGD puts more emphasis on the 2-bit parameter loss function (see Fig.~\ref{fig: ablation}(a)), while ignoring the 3-bit and 4-bit loss and further aggravates the imbalance between loss functions.
Beyond these points, it slows down the overall training speed since MGD needs to solve the problem in Eq.~\eqref{eq: mgd} once per iteration. Therefore, the US strategy is the best choice regarding performance and speed.

\subsubsection{Effect of Self-Knowledge Distillation}
\label{subsubsec: ablation self-KD} 

We further study the effectiveness of the self-KD paradigm described in Sec.~\ref{subsubsec: self-knowledge distillation}. 
Table~\ref{tab: ablation self-KD} compares the training results with (\textit{w.}) and without (\textit{w/o}) self-KD. 
As an effortless and lightweight training technique, self-KD positively affects the performance of vertical-layered models, especially for low-bit-width networks.
In particular, the self-KD improves the 2-bit ResNet-50 by $0.8\%$ (from 72.4\% to 73.2\%) in Top-1 accuracy. 
We analyze the quantization error of activations of the last five convolution layers from well-trained ResNet-50 models running at a 2-bit setting in Fig.~\ref{fig: ablation}(c). 
It is analyzed that deeper layers are more prone to activation quantization, which significantly degrades the network performance \cite{Characterization_Boo}.
With the self-KD paradigm here, the averaged quantization errors of activation in the last two layers are dramatically reduced compared with the setting without self-KD, hence improving the quantized model performance by feeding more accurate features into the final classifier.

\begin{table}[h]
\renewcommand\arraystretch{1.4}
    \centering
    \caption{Experiment results of using different self-KD settings in joint training are presented. 
    We use KL and CS to denote the Kullback-Leibler divergence loss and the cosine similarity loss respectively. } 
    \begin{small}
    \setlength{\tabcolsep}{4.5mm}{
    \begin{tabular}{c|c|ccc}
    \bottomrule[1pt]
    \multirow{2}{*}{Model} & \multirow{2}{*}{Self-KD} & \multicolumn{3}{c}{Top-1 Acc (\%) @ W/A} \\ \cline{3-5}  &      & 2 / 2  & 3 / 3  & 4 / 4   \\ \clineB{1-5}{2.5}
    \multirow{3}{*}{ResNet-18}  & \textit{w/o}  & 66.7   & 69.9   & 70.6    \\
    						  & KL            & 66.7   & 69.9   & 70.4    \\ 
                                & CS            & \textbf{67.1}   & \textbf{70.1}   & \textbf{70.6}    \\ \cline{1-5}
    \multirow{3}{*}{ResNet-50}  & \textit{w/o}  & 72.4   & 75.9   & \textbf{76.8}    \\
						    & KL            & 72.1   & 75.8   & \textbf{76.8}    \\ 
                    		& CS            & \textbf{73.2}   & \textbf{76.1}   & \textbf{76.8}    \\ 
    \toprule[1pt]
    \end{tabular}
    }
    \end{small}
    \label{tab: ablation self-KD}
\end{table}

Moreover, we also study the selection of the self-KD loss function by comparing the cosine similarity (CS) based self-KD with the Kullback–Leibler-divergence (KL) based one in Table~\ref{tab: ablation self-KD}. 
The experimental results show that the selection of distillation loss function is crucial, and the KL loss is almost useless. 
This is because the classic KL distillation loss may mislead model training since the teacher is sometimes unreliable, while the cosine similarity function serves as a more robust choice for ambiguous knowledge \cite{SKDQ_Boo}.

\subsubsection{Effects of Replacing Full Precision First and Final Layers}
\label{subsubsec: ablation FP} 

For the quantization of neural networks, it is a common practice to keep the first convolutional layer and the last fully connected layer in full precision to preserve accuracy \cite{XOR_Rast}. 
Our vertical-layered models share these parameters among quantized networks of different precision for convenient deployments. 
One might wonder whether the quantized networks' performance can be enhanced by making the first and last layers tailored to different bit widths. 
Under this circumstance, when the devices require more enhance layers, the server must also offload the corresponding first and last layers.
We experiment with ResNet-18 and ResNet-50 trained on ImageNet, and the result is shown in Table~\ref{table: ablation FP}. 
We can observe that replacing these two layers shows little effect on the accuracy improvement at the cost of a large portion of overhead parameters in ResNet-18. 
Regarding the ResNet-50 model, the results demonstrate that the joint optimization of parameters in these two layers, \ie using the same layers for different bit-width networks in Eq.~\eqref{eq: unitary scalarization}, assists the quantized model training, especially for the lowest 2-bit quantized one.  
Hence, we do not recommend training individual first and last layers for scalable bit-width inference.

\subsection{Mixed-Precision Network}
\label{sec: mixed precision}

\begin{figure}[h]
    \centering
    \includegraphics[width=0.95\linewidth]{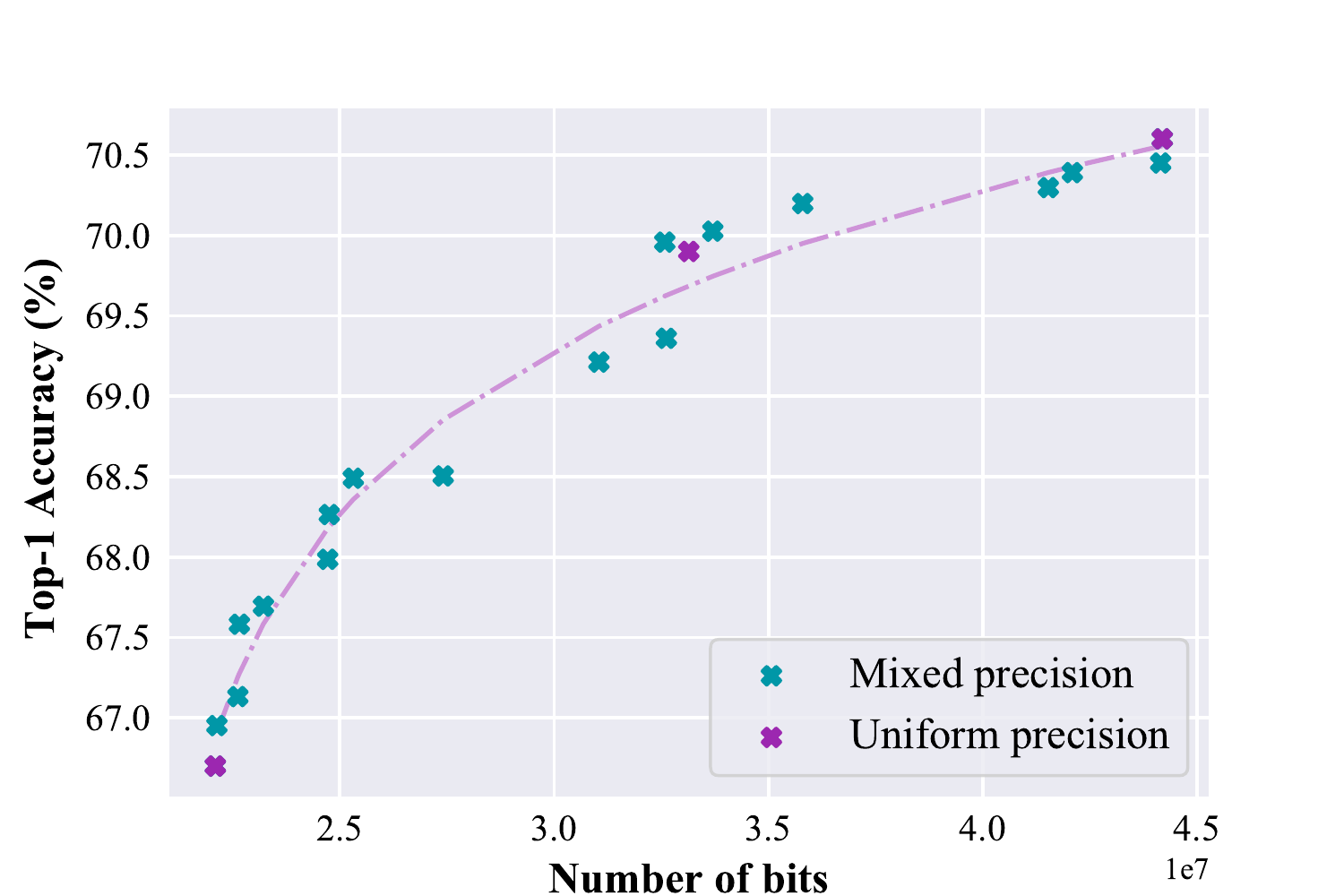}
    \vspace{-0.2cm}
    \caption{Top-1 accuracy vs. model size for mixed precision quantized ResNet-18 on ImageNet dataset.}
    \label{fig: mixed precision}
\end{figure}

In this part, we assess whether our vertical layered model supports the mixed-precision setting well. 
We accommodate our quantization network to the mixed-precision scenario according to the criterion of minimizing the quantization error given a limited number of bits on model weights at the unit of each layer, see Eq.~\eqref{eq: mixed precision sampling}. We experiment with ResNet-18 on ImageNet and plot the network accuracy against the respective model size in Fig.~\ref{fig: mixed precision}. Since the BN statistics vary in different precisions, the BN statistics in the sampled network are re-estimated using the training data. We can observe that the Top-1 validation accuracy increases with the enlarging of the overall number of bits in general, manifesting the effectiveness of our model in the mixed-precision setting. 
We note that the 3-bit uniform precision network performs worse than the mixed-precision quantized one with fewer bits requirement. 
This phenomenon can be explained by the different sensitivity for quantization over all layers in one network model \cite{HAWQ_Dong}, which highlights the advantages of mixed-precision quantization. 

\section{Conclusion}
\label{Sec: conclusion}

We have proposed a vertical-layered representation for quantizing deep neural networks, which allows heterogeneous devices to access the quantized model in an on-demand manner for scalable and efficient inference. 
To obtain a high-performance model that supports vertical-layered representation, we propose a once quantization-aware training framework with a downsampling operation, and multi-objective optimization techniques are employed to balance the collaborative training process. 
Further, we find that the self-knowledge distillation paradigm can boost the training of the low-bit-width networks with the proper selection of a distillation function.
Extensive experimental results demonstrate the feasibility of vertical-layered representation of quantized networks, where our model delivers competitive performance in all assessed precision settings.
Our vertical-layered model achieves significant storage and transmission cost compression for scalable inference. 
We hope our study would inspire more research investigations on employing vertical-layered weight representation for quantization in heterogeneous application scenarios.


\bibliographystyle{IEEEtran}
\bibliography{IEEEabrv,egbib}

\end{document}